\documentclass[10pt,twocolumn,letterpaper]{article}

\usepackage{cvpr}

\definecolor{cvprblue}{rgb}{0.21,0.49,0.74}
\usepackage[table]{xcolor}
\definecolor{agent}{HTML}{CB6D51}
\definecolor{environment}{HTML}{7bb09b}
\definecolor{object}{HTML}{4D419A}
\usepackage[pagebackref,breaklinks,colorlinks,allcolors=cvprblue]{hyperref}

\begin{document}

\title{\LARGE Towards Spatial Supersensing in the Wild} 

\author{
  Tianjun Gu\textsuperscript{1}\thanks{Equal contribution.} \quad
  Tianyu Xin\textsuperscript{1}\footnotemark[1] \quad
  Kuan Zhang\textsuperscript{1}\footnotemark[1] \quad
  Bowen Yang\textsuperscript{1} \quad
  Kok-Chung Chua\textsuperscript{1} \\
  Peize Li\textsuperscript{1} \quad
  Xinran Zhang\textsuperscript{1} \quad
  Yupeng Chen\textsuperscript{1} \quad
  Qiyue Zhao\textsuperscript{1} \quad
  Qinlei Xie\textsuperscript{1} \\
  Jianhang Liu\textsuperscript{1} \quad
  Yucheng Lu\textsuperscript{1} \quad
  Yinan Han\textsuperscript{1} \quad
  Marco Pavone\textsuperscript{2,3} \quad
  Yiming Li\textsuperscript{1}\footnotemark[2] \vspace{2mm} \\ 
  $^{1}$Tsinghua University \quad 
  $^{2}$NVIDIA \quad 
  $^{3}$Stanford University \\
  {\small \url{https://vsi-super-wild.github.io}}
  \vspace{-10mm}
}

\twocolumn[{
\renewcommand\twocolumn[1][]{#1}
\vspace*{-9mm}
\maketitle

\begin{center}
    \captionsetup{type=figure}
    \includegraphics[width=0.77\textwidth]{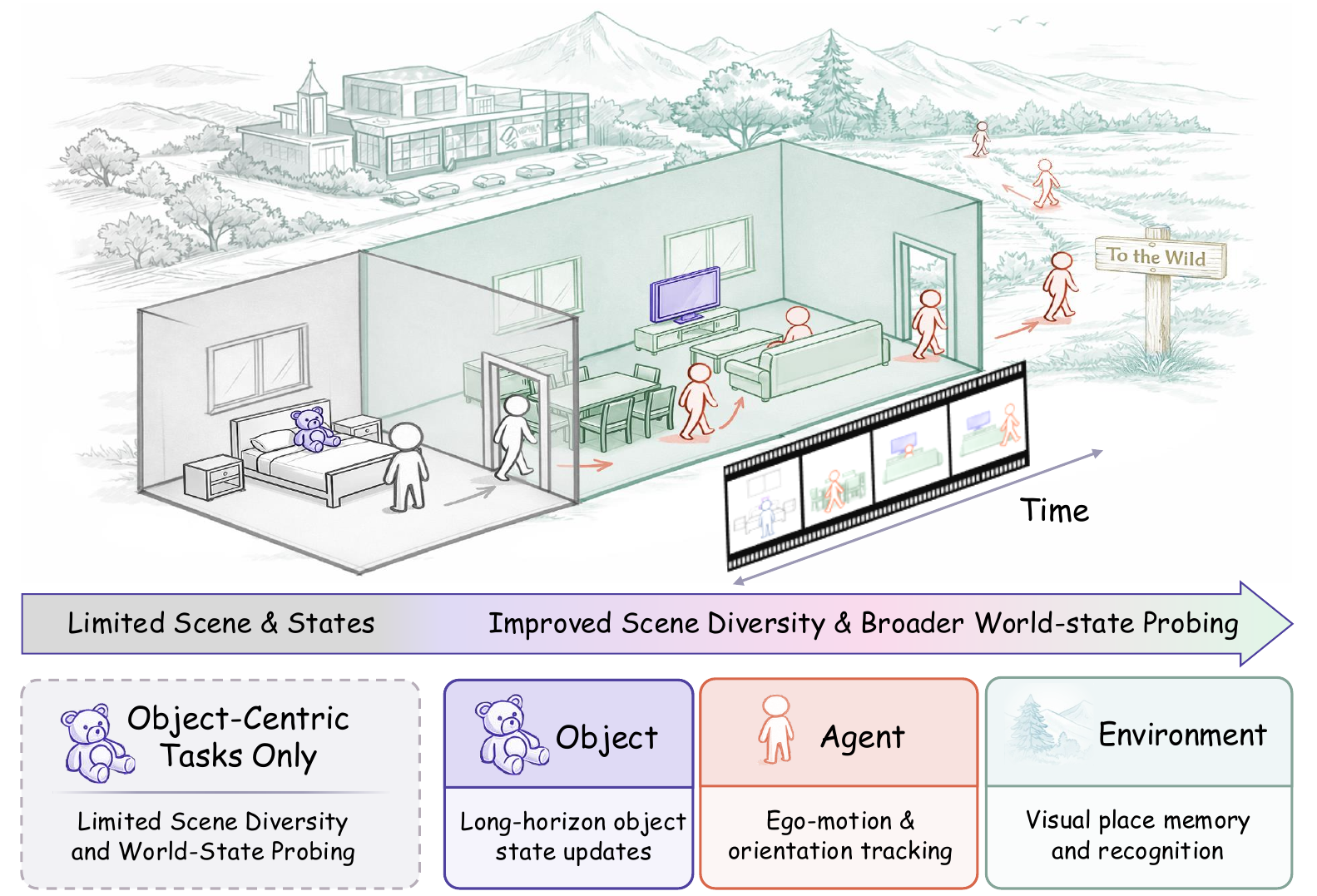}
    \vspace{-5pt}
    \captionsetup{width=0.79\textwidth}
    \caption{We advance spatial supersensing beyond household object modeling to world modeling across three anchors—agent, object, environment—using long-form, in-the-wild videos from diverse real-world scenes.}
    \label{fig:teaser}
\end{center}

}]

{
\renewcommand{\thefootnote}{}
\footnotetext{* Equal contribution. \quad † Corresponding author.}
}

\begin{abstract}
Humans can efficiently parse continuous sensory streams, from hours to years, scaffolding an internal world model that grounds spatial reasoning and prediction. To mimic this capacity, spatial supersensing~\cite{yang2025cambrians} challenges multimodal models to move beyond linguistic understanding toward true world modeling. However, their benchmark relies on synthetic long videos—formed by concatenating random short clips—and is mostly limited to household scenes, leaving real-world continuity and diversity underexplored. To address the gap, we introduce \textbf{\textsc{VSI-Super-Wild}}, a large-scale benchmark for evaluating spatial supersensing over long temporal horizons in diverse in-the-wild scenes. Notably, inspired by cognitive studies on how humans structure experience, we systematically probe the full triad of world state: the agent (observer), objects (scene items), and the environment (places and global layout). In total, \textsc{VSI-Super-Wild} contains \textbf{6,980} human-verified question-answer pairs derived from \textbf{442} real-world videos spanning 8 scene categories, including long-form recordings exceeding 4 hours. Results on \textsc{VSI-Super-Wild} expose a fundamental disconnect: despite advances in static image understanding, models consistently fail at tasks that require coherent world-state tracking over time. We characterize how performance degrades with world-state complexity and temporal horizon, and diagnose four failure modes: spatial collapse, semantic shortcuts, insufficient update, and instance confusion. This taxonomy reveals that models lack mechanisms to bind objects, agents, and environments into a unified spatial world model—a fundamental gap that defines the path forward for spatial supersensing. \hspace{-5pt}
\end{abstract}

\section{Introduction}
Humans bind what we see (\eg, objects and places) with what we do (\eg, ego motion and actions) into a unified world model from continuous visual experience, enabling us to answer, days or months later, where we left our keys or whether we turned off the stove~\cite{autobiomemory2011, autobiomemory2024}. Cognitive science suggests that this capability does not arise from \emph{recording a complete, verbatim stream of visual input}, but from \emph{constructing and maintaining internal representations} that integrate external information (\eg, object features and environmental geometry) with self-related information (\eg, ego pose and motion), thereby modeling how the world evolves over space and time \cite{rensink1997tosee, burgess2006egocentricallocentric, rememberpastandimaginefuture, Bainbridge2017Memorability}.

How close are existing multimodal large language models (MLLMs) to building internal world models from streaming visual inputs? Despite rapid progress on video benchmarks, many tasks can still be solved with sparse-frame evidence, local matching, or textual priors—leaving true world modeling unexplored~\cite{videomme, mangalam2023egoschemadiagnosticbenchmarklongform, li2024mvbenchcomprehensivemultimodalvideo, videommmu, brown2025vsi_debiased}. To push beyond language-only understanding, Cambrian-S introduces \emph{spatial supersensing}~\cite{yang2025cambrians}: the ability of an MLLM to construct, update, and predict with an implicit 3D world model from continual sensory experience. Its benchmark, \textsc{VSI-Super}, defines two challenging tasks, \ie, spatial recall (VSR) and continual counting (VSC), to test whether MLLMs can construct and maintain recallable world states across long horizons and changing viewpoints.

Despite its pioneering contributions, \textsc{VSI-Super} leaves several key dimensions of spatial supersensing underexplored. \emph{First}, it predominantly includes household scenes, with limited exposure to in-the-wild diversity~\cite{yu2025thinking360deghumanoidvisual}. \emph{Second}, its long videos are synthesized by concatenating random short clips with in-frame editing, departing from fully natural sensory streams. \emph{Third}, it emphasizes object-centric probing while underrepresenting agent- and environment-centered aspects of world state. These limitations motivate a new benchmark moving spatial supersensing toward real streams, real scenes, and full world-state coverage.

To bridge these gaps, we introduce \textbf{\textsc{VSI-Super-Wild}}, a large-scale video benchmark for evaluating spatial supersensing in the wild. It contains 6,980 human-verified Q\&A pairs derived from 442 real-world videos, including long-form recordings exceeding 4 hours. As illustrated in Fig.~\ref{fig:teaser}, \textsc{VSI-Super-Wild} advances beyond prior work along two axes: \emph{real-world diversity} and \emph{world-state coverage}.
\begin{itemize}
    \item For \emph{real-world diversity}, we curate long-form in-the-wild videos spanning diverse scene categories, camera movements, and object semantics—with natural continuity and no generative editing or clip concatenation.
    \item For \emph{world-state coverage}, we broaden evaluation beyond objects through a cognitively grounded, multi-anchor design inspired by neuronal vector coding~\cite{BicanskiBurgess2020NeuronalVectorCoding}. We view spatiotemporal world states as organized around three~complementary anchors—\emph{agent}, \emph{objects}, and \emph{environment}.
\end{itemize}
Specifically, \textsc{VSI-Super-Wild} includes four tasks that target different anchors of world state: \textbf{VMR} (motion orientation recall) probes the \emph{agent} by asking questions about the camera motion orientation; \textbf{VPO} (place temporal ordering) probes the \emph{environment} by testing memory of place order under varying viewpoints; \textbf{VOO} (object temporal ordering) probes \emph{objects} by requiring temporal reasoning about when specific items appear; and \textbf{VOC} (continuous object counting) integrates across anchors by requiring persistent tracking of objects as the agent moves through the environment. Together, these tasks require true world modeling that can unify agent, object, and environment across space and time.

We evaluate 13 mainstream MLLMs on \textsc{VSI-Super-Wild}, where they exhibit consistently weak performance—often near chance--and suggest they still struggle to \emph{maintain coherent, recallable world states} under the real, diverse in-the-wild settings.
To pinpoint where this gap is most pronounced, we conduct \emph{quantitative analyses on task demands and temporal horizon}.
Across tasks, models are relatively more reliable on object-centric state queries, but degrade noticeably on queries that require stronger 3D spatial cognition about the environment or agent.
Across temporal horizons, performance drops substantially as videos become longer, indicating difficulty in sustaining world-state updates over extended streams and a tendency for errors to accumulate rather than be corrected with additional evidence.
We further conduct \emph{qualitative error analyses} to understand why these failures occur, and observe recurring patterns of spatial collapse, semantic shortcuts, insufficient update, and instance confusion, reinforcing that today’s MLLMs often default to frame-level semantics and lack stable spatiotemporal representations that support long-horizon reasoning.
Taken together, these observations suggest that progress on spatial supersensing in the wild may hinge on more principled spatiotemporal world modeling—i.e., representations that move beyond frame-level semantics while remaining consistent and coherent under long-horizon, continually updated streams.

In summary, our contributions are:
\begin{itemize}
    \item \textbf{In-the-Wild Video Benchmark.} We curate 442 videos across 8 scene categories, with 6,980 human-verified Q\&As, providing a large-scale benchmark for spatial supersensing in unconstrained and real-world settings.
	\item \textbf{Multi-Anchor Task Suite.} We design four cognitively grounded tasks that probe world states spanning agent, object, and environment anchors to systematically evaluate implicit world modeling over long temporal horizons.
	\item \textbf{Diagnostic Insights.} We benchmark thirteen mainstream MLLMs on \textbf{\textsc{VSI-Super-Wild}}, delivering task- and horizon-wise quantitative analyses and qualitative error diagnoses that expose recurring failure modes and identify open challenges for spatial supersensing.
\end{itemize}

\section{Related Work}

\noindent \paragraph{Video Multimodal Large Language Models.}
Recent Multimodal Large Language Models (MLLMs) signify a strategic paradigm shift from static image comprehension \cite{alayrac2022flamingo, li2023blip2, liu2023llava, zhu2024minigpt4, dai2023instructblip} to complex video streams understanding \cite{llava-video, llava-onevision, vila, video-R1, apollo}.
This modality shift from image to video naturally introduces \emph{temporality}, creating growing demands for memory and streaming mechanisms that can retain, update, and process information over extended visual sequences \cite{zhang2025flashvstream, qian2025dispider, di2025rekv, simplestream2026, streamchat2026}.
Organized as sequences of 2D frames, videos also push models beyond frame-wise understanding and infer an underlying 3D spatial world, motivating recent efforts on spatial understanding and reasoning \cite{spatialmllm, mindjourney, spatiallm, sensenova-si, Chen2024spatialvlm, cheng2024spatialrgpt, ma2025spatialreasoner}.
Building on progress in temporal modeling and spatial reasoning, several recent works have begun to further explore whether MLLMs can \emph{model the world} by constructing and updating the world states over continuous visual experience \cite{yang2025cambrians, spatialttt2026}.
However, current validation of such world-modeling ability remains underexplored on several key dimensions with simplified evaluation settings, leaving the need for more rigorous evaluation of how MLLMs construct and update spatiotemporal world states from real-world continuous visual experience.

\noindent \paragraph{Video Understanding Benchmarks for MLLMs.}
Existing video benchmarks for MLLMs already span a broad landscape, including general video understanding \cite{videomme, videommmu, tomato, li2024mvbenchcomprehensivemultimodalvideo, perception}, long-horizon video evaluation \cite{hourvideo, Longvideobench, LVbench2025}, and 3D spatial reasoning \cite{VSI-bench, mindcube, ramakrishnan2025space, yeh2026allangles, li2025stibench, xu2025multispatialmllm, gu2026visionlanguagemodelslaghuman}.
However, empirical results show that they still highly rely on linguistic priors \cite{language_shortcut_1, language_shortcut_2, language_shortcut_3}, and often answer QAs correctly by semantic shortcut rather than establishing a coherent world representation \cite{brown2025vsi_debiased}.
\textsc{VSI-Super} therefore marks an important step by evaluating the \emph{spatial supersensing} capability of MLLMs, which explicitly aims to reduce such shortcuts and push evaluation toward implicit 3D world modeling and predictive world-state understanding \cite{yang2025cambrians}.
Yet several important dimensions remain underexplored, as its evaluation primarily focuses on object-centric states, while its data are built from relatively synthetic and scene-restricted settings rather than diverse real-world continuous videos.
These limitations motivate the need for a more rigorous benchmark that evaluates MLLM world modeling over broader world-state coverage and more diverse in-the-wild scenes.

\noindent \paragraph{World Modeling over Video Streams.}
World models, broadly understood as neural systems that internalize and simulate the spatiotemporal dynamics of the physical world, have recently gained renewed attention in both generative paradigms \cite{worldlabs2025marble, sora, mindjourney, genie3} and representation-learning paradigms \cite{V-JEPA, assran2025vjepa2, murlabadia2026vjepa2_1, leworldmodel2026}.
In parallel, large-scale pretraining could also endow MLLMs with knowledges about spatiotemporal dynamics of the real world, and yet their capability of world modeling has not yet been fully examined.
Cambrian-S~\cite{yang2025cambrians} takes an important step in this direction by introducing MLLM-based predictive world modeling, where models must construct and update world states over continuous video streams. 
Nevertheless, several important dimensions remain underexplored, as it primarily centers on object-level state modeling and synthetic long-video construction.
This raises a natural question: how should MLLM-based world modeling be evaluated more comprehensively?
Theories in cognitive science, such as neuronal vector coding \cite{BicanskiBurgess2020NeuronalVectorCoding}, suggest that biological spatiotemporal cognition is organized around three anchors: \emph{the agent, the object, and the environment}. This perspective motivates the evaluation of world modeling in MLLMs with richer world states and more diverse real-world scenes.

\section{\textsc{VSI-Super-Wild}}
\label{sec:VSI_SUPER_Wild}

\begin{figure*}[t]
    \centering
    \includegraphics[width=\linewidth]{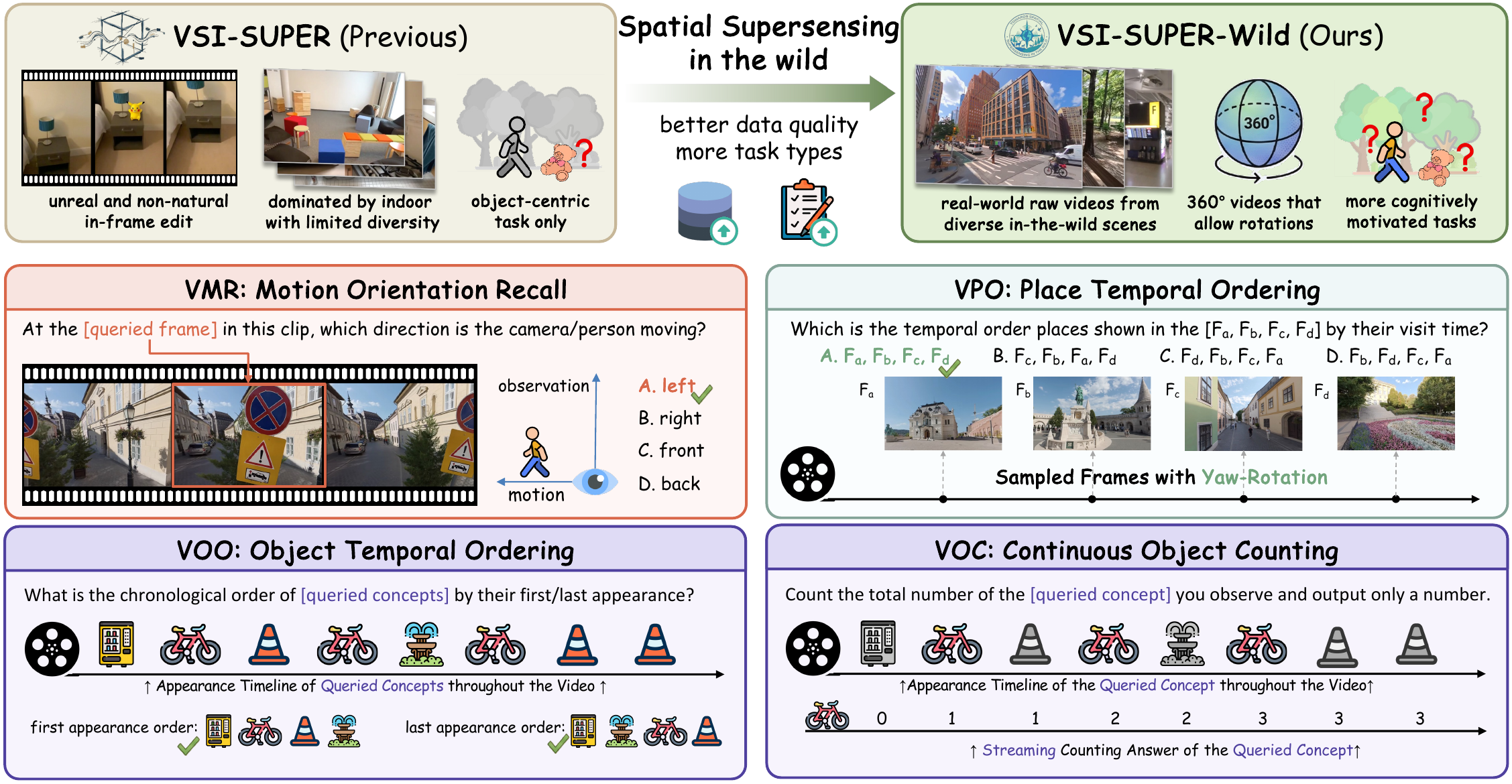}
    \caption{Overview of \textsc{VSI-Super-Wild}. We extend spatial supersensing from synthetic indoor scenes to real-world, 360° outdoor environments with better data quality and more task types. Our benchmark probes the agent-object-environment triad through four tasks: \textbf{VMR} (motion orientation recall), \textbf{VPO} (place temporal ordering), \textbf{VOO} (object temporal ordering), and \textbf{VOC} (continuous object counting). Together, these tasks demand coherent world modeling across diverse, unconstrained scenarios.}
    \label{fig:task_illustration}
\end{figure*}

To bridge the gap between existing benchmarks and the real-world spatial supersensing, we introduce \textsc{VSI‑Super‑Wild}, constructed from genuinely long‑form, in‑the‑wild videos. This section details the design and construction of our benchmark. We first motivate the four tasks (§\ref{subsec:task_mot_setup}) through the lens of cognitive science. We then describe the semi‑automatic data curation pipeline with human‑in‑the‑loop verification (§\ref{subsec:data_construction}) and present comprehensive dataset statistics that highlight the scale, diversity, and complexity of \textsc{VSI‑Super‑Wild} (§\ref{subsec:bench_statis}).

\subsection{Task Motivation and Setup}

\label{subsec:task_mot_setup}

To evaluate spatial supersensing in unconstrained settings, \textsc{VSI-Super-Wild} is designed around the triad of world modeling: \emph{the {\color{agent}agent}, {\color{object}objects}, and the {\color{environment}environment}.} By leveraging genuinely long-form, unedited panoramic videos and comprehensive tasks, our benchmark enables broader world state probing and improved real-world diversity compared to prior benchmarks. We define four cognitively grounded tasks to systematically assess how well MLLMs can construct and maintain these specific world states.

\noindent \paragraph{V{\color{agent}MR}: {\color{agent}M}otion Orientation {\color{agent}R}ecall.}
After observing a long video, \textbf{VMR} requires MLLMs to infer the motion orientation relative to its viewing direction at a queried moment, which is specified by the frame sampled there. Such task \emph{cannot} be solved by frame-level matching on the queried moment, since a single frame is often insufficient to determine motion orientation. Instead, the model must recall the latent world state at that moment to decode the corresponding motion orientation. The VMR benchmark therefore probes spatial supersensing by \emph{requiring implicit world modeling} over long-horizon observations, without which such implicit motion states cannot be reliably recalled.

\noindent \paragraph{V{\color{environment}PO}: {\color{environment}P}lace Temporal {\color{environment}O}rdering.}

After observing a long panoramic video, VPO asks the model to determine the temporal order of four query frames sampled from different moments representing different places. Each query frame is yaw-rotated to a different viewing direction, preventing trivial frame matching solution. We construct VPO benchmark from panoramic videos so that these rotations are lossless and realistic. With the same place presented under different headings, VPO thus probes spatial supersensing by testing whether models can maintain implicit, heading-invariant place representations and use them to reconstruct the observer’s experience over space and time.

\noindent \paragraph{V{\color{object}OO}: {\color{object}O}bject Temporal {\color{object}O}rdering.}
After observing a long video, VOO queries a set of objects and asks the model to output their temporal order conditioned on the queried occurrence type, i.e., the \emph{first} or \emph{last} time each object becomes visible. Compared to the VSR benchmark in VSI-SUPER, VOO differs in two key aspects: (i) it uses real, unedited object observations, whereas VSR relies on generative, non-natural objects inserted via in-frame editing; and (ii) it explicitly requires distinguishing first-versus-last occurrences of queried objects within a continuous stream. Consequently, VOO probes spatial supersensing by testing whether models can perform \emph{self-updating} world modeling—maintaining a dynamically updated world state of object encounters over time and space to support occurrence-conditioned queries.

\noindent \paragraph{V{\color{object}OC}: Continuous {\color{object}O}bject {\color{object}C}ounting.}
After observing a long video stream, VOC queries an object category and requires the model to output its unique-instance counting number. This task follows the VSC setup in VSI-SUPER, but is instantiated on real, unedited long in-the-wild videos, where the visual evidence for counting is more natural and video streams are more coherent. Consequently, VOC probes spatial supersensing by testing whether models can perform long-horizon, \emph{self-updating} world modeling that maintains a consistent latent state of object counts over time and grounding it into an accurate numerical answer.

\begin{figure*}[t]
    \centering
    \includegraphics[width=0.9\linewidth]{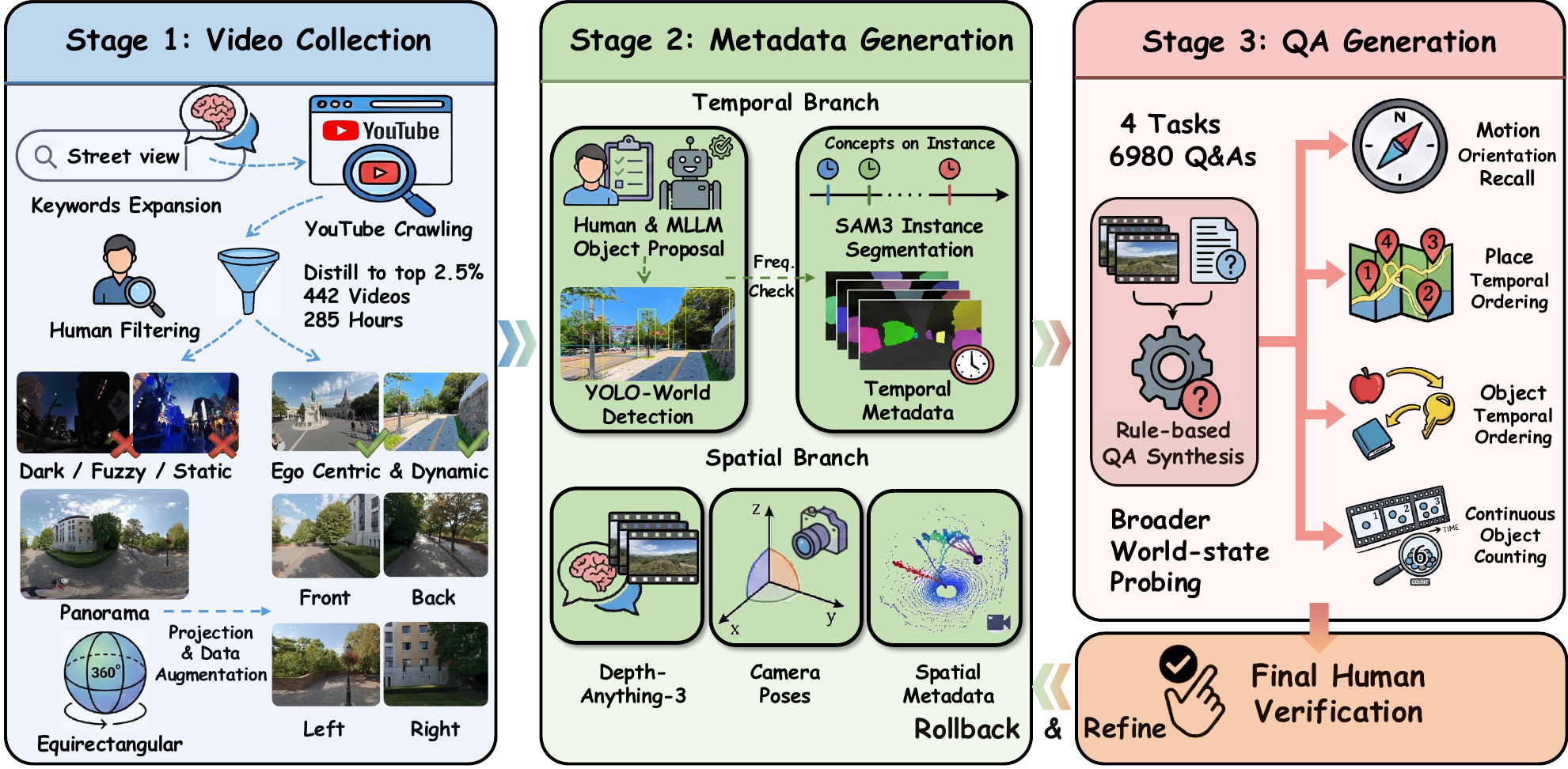}
    \caption{Data construction pipeline of \textsc{VSI-Super-Wild}. We design a semi-automatic pipeline including video collection, metadata generation, Q\&A generation, with human-in-the-loop verification. }
    \label{fig:data_construction}
\end{figure*}

\subsection{Dataset Construction}

\label{subsec:data_construction}

As shown in Fig.~\ref{fig:data_construction}, the construction of our dataset follows a \emph{scalable}, \emph{semi-automatic} pipeline with \emph{human-in-the-loop} verification, consisting of three stages: {video collection}, {metadata generation}, and {question-answer generation}. We describe each stage in detail below.

\noindent \paragraph{Video Collection.}
We crawl in-the-wild panoramic videos from YouTube spanning eight scene categories (culture \& entertainment, industry, medical, office \& education, residential space, retail space, street view, and transportation hubs). To ensure data quality, we manually filter the crawled videos and retain predominantly egocentric and dynamic recordings. As panoramic videos are typically stored in an equirectangular format, we project each panorama into four orthogonal perspective views (Front/Back/Left/Right) and treat perspective videos as independent inputs for subsequent processing. Unless otherwise specified, the term ``video'' hereafter refers to these perspective videos.

\noindent \paragraph{Metadata Generation.}
From each video, we generate temporal and spatial metadata to support downstream Q\&A construction. First, \emph{temporal metadata} captures object appearances over time by (i) video-specific candidate objects proposed by human experts and an MLLM, (ii) filtering candidates with YOLO-World\cite{yoloworld} based on their temporal occurrence statistics, and (iii) applying Segment-Anything-3 (SAM3)\cite{sam3} to produce timestamped instance masks for the retained objects. Second, \emph{spatial metadata} consists of time-varying camera poses estimated across the video using Depth-Anything-3 (DA3)\cite{depthanything3}.

\noindent \paragraph{Question-Answer Generation.}
Given the videos and their temporal and spatial metadata, we synthesize Question–Answer (Q\&A) pairs for four tasks: \textbf{VMR} (motion orientation recall), \textbf{VOO} (object ordering), \textbf{VPO} (place ordering), and \textbf{VOC} (object counting), where the prefix ``V'' denotes \textsc{VSI-Super-Wild}. Our Q\&A construction is rule-based, with answers provided in either \emph{multiple-choice (MC)} or \emph{numeric} formats. Finally, we incorporate human verification to ensure Q\&A correctness, rolling back to refine metadata or Q\&A when necessary.

\subsection{Benchmark Statistics}

\label{subsec:bench_statis}

\begin{figure*}[t]
\includegraphics[width = \textwidth]{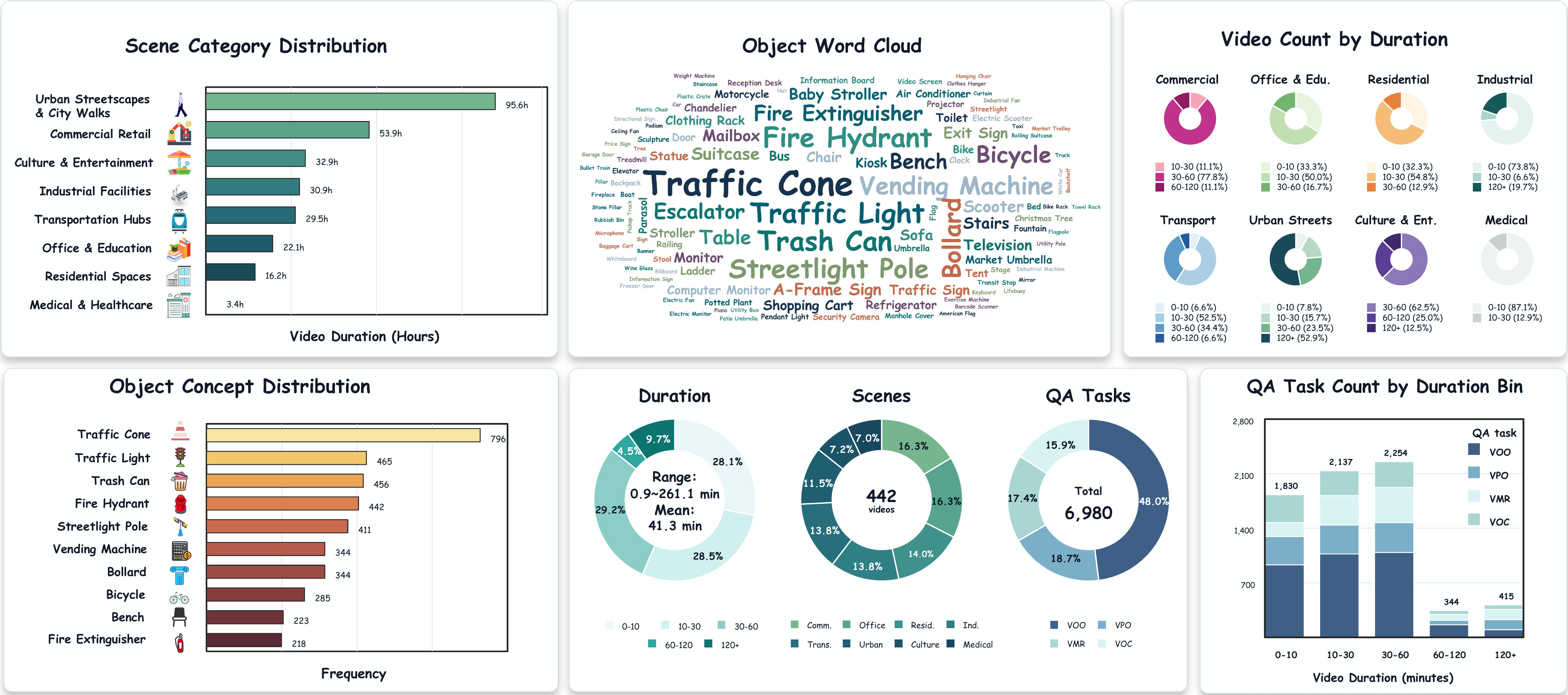}
\vspace{-3mm}
\caption{Statistics of \textsc{VSI-Super-Wild}. The figure reports scene-category duration statistics, object-frequency statistics from VOO/VOC tasks, video-count distributions over duration bins for each scene category, object-concept frequencies, overall single-video duration and QA task distributions, and QA task counts by duration bin.}
\vspace{-8mm}
\label{benchstats}
\end{figure*}

After we detail the rigorous curation and synthesis pipeline, the intrinsic value of \textsc{VSI-Super-Wild} lies in the empirical complexity, multi-dimensional distribution, and cognitive difficulty of the video data and tasks. In this section, to demonstrate the scale, diversity, and complexity of \textsc{VSI-Super-Wild}, we present a statistical analysis of the \textsc{VSI-Super-Wild} dataset (see Fig.~\ref{benchstats}). We deconstruct the data to provide a more comprehensive understanding, including video \& meta information, question-answer pairs. 

\noindent \paragraph{Video \& Meta Information.}
In total, \textsc{VSI-Super-Wild} collects 442 high-quality view-level videos, yielding 284.52 hours (17,071.32 minutes) of egocentric experience in the wild. To ensure that the benchmark accurately reflects open-world complexity, the videos span 8 distinct scene categories. Commercial retail accounts for 16.29\% of videos by count and 18.95\% by duration, while transportation hubs account for 13.80\% by count and 10.38\% by duration. Unlike highly edited indoor clips, our videos have an average duration of 38.62 minutes (ranging from 0.87 to 261.08 minutes), preserving the natural temporal continuity required for implicit 3D world modeling.

\noindent \paragraph{Question-Answer Pairs.}
Our pipeline yielded a total of 6,980 high-quality Q\&A pairs. To ensure a comprehensive evaluation of the agent-object-environment triad, the questions are distributed across the four proposed tasks: VMR (1,215 pairs), VPO (1,302 pairs), VOO (3,350 pairs), and VOC (1,113 pairs). The VOO split contains two balanced subtypes, with 1,675 first-appearance recall questions and 1,675 last-appearance recall questions. Furthermore, we strictly controlled the answer distributions to prevent models from exploiting statistical shortcuts.

\section{Experiment}
\definecolor{proprietaryheader}{HTML}{FFF0F5} 
\definecolor{opensourceheader}{HTML}{F0F8FF} 

\begin{table*}[t]
\scriptsize
\caption{\textbf{Performance on \textsc{VSI-Super-Wild}.} We report task scores as mean \(\pm\) standard error of the mean (SEM) for \textbf{VMR}, \textbf{VPO}, \textbf{VOO} (accuracy, \%), and \textbf{VOC} (MRA, \%). \textbf{Overall} denotes the question-level aggregate score across all tasks. (\textbf{Bold}: best, \underline{underline}: second best)}
\label{tab:main-exp}
\vspace{-4pt}
\setlength{\tabcolsep}{5pt}
\renewcommand\arraystretch{1.15}
\resizebox{\textwidth}{!}{%

\begin{tabular}{llccccc}
\toprule
\textbf{Model} & \textbf{Base LM} & \textbf{VMR} & \textbf{VPO} & \textbf{VOO} & \textbf{VOC} & \textbf{Overall} \\
\midrule
\rowcolor{proprietaryheader}\multicolumn{7}{l}{\textit{Proprietary Models}} \\
GPT-5.4 & UNK. & \textbf{37.76\(\pm\)3.02} & 24.87\(\pm\)2.63 & 37.62\(\pm\)3.13 & 32.94\(\pm\)1.11 & \underline{34.52\(\pm\)1.68} \\
Gemini-3.1-Pro & UNK. & \underline{34.21\(\pm\)2.26} & \textbf{63.84\(\pm\)4.46} & \textbf{42.54\(\pm\)1.35} & \textbf{38.16\(\pm\)5.13} & \textbf{44.36\(\pm\)1.39} \\
Gemini-3.1-FlashLite & UNK. & 28.96\(\pm\)2.30 & 23.81\(\pm\)1.75 & 23.17\(\pm\)0.85 & 20.35\(\pm\)1.80 & 23.85\(\pm\)0.72 \\
\midrule
\rowcolor{opensourceheader}\multicolumn{7}{l}{\textit{Open-Source Models}} \\
Cambrian-S-0.5B & Qwen2.5-0.5B & 24.36\(\pm\)1.23 & 10.14\(\pm\)0.84 & 20.87\(\pm\)0.70 & 33.30\(\pm\)0.95 & 21.46\(\pm\)0.46 \\
Cambrian-S-3B & Qwen2.5-3B & 25.68\(\pm\)1.25 & 25.42\(\pm\)1.21 & 38.63\(\pm\)0.84 & 34.06\(\pm\)0.96 & 33.18\(\pm\)0.54 \\
Cambrian-S-7B & Qwen2.5-7B & 26.91\(\pm\)1.27 & 25.58\(\pm\)1.21 & 40.36\(\pm\)0.85 & 33.81\(\pm\)0.97 & 34.22\(\pm\)0.54 \\
Cambrian-S-7B-LFP$^{*}$ & Qwen2.5-7B & 28.62\(\pm\)1.20 & 26.06\(\pm\)2.46 & 39.26\(\pm\)3.69 & 26.17\(\pm\)7.98 & 32.86\(\pm\)2.24 \\
Cambrian-S-7B-LFP & Qwen2.5-7B & 26.58\(\pm\)1.27 & 26.11\(\pm\)1.22 & 40.27\(\pm\)0.85 & 28.36\(\pm\)0.94 & 33.35\(\pm\)0.54 \\
InternVL3.5-8B & Qwen3-8B & 27.90\(\pm\)1.29 & 24.65\(\pm\)1.19 & 35.13\(\pm\)0.82 & \underline{36.74\(\pm\)0.96} & 32.18\(\pm\)0.53 \\
Qwen2-VL-7B & Qwen2-7B & 27.41\(\pm\)1.28 & 23.66\(\pm\)1.18 & 30.21\(\pm\)0.79 & 18.72\(\pm\)7.98 & 26.67\(\pm\)1.37 \\
Qwen2.5-VL-7B & Qwen2.5-7B & 31.52\(\pm\)1.33 & \underline{26.11\(\pm\)1.22} & 35.61\(\pm\)0.83 & 22.11\(\pm\)0.87 & 30.98\(\pm\)0.53 \\
Qwen3-VL-8B & Qwen3-8B & 25.60\(\pm\)1.25 & 24.35\(\pm\)1.19 & 40.36\(\pm\)0.85 & 32.73\(\pm\)7.98 & 33.59\(\pm\)1.37 \\
Qwen3.5-9B & Qwen3.5-9B & 25.68\(\pm\)1.25 & 25.19\(\pm\)1.20 & \underline{41.25\(\pm\)0.85} & 30.76\(\pm\)0.95 & 33.87\(\pm\)0.54 \\
Spatial-TTT-nano$^{*}$ & Qwen3-2B & 24.53\(\pm\)1.23 & 24.12\(\pm\)1.19 & 27.82\(\pm\)0.77 & 35.93\(\pm\)0.96 & 27.85\(\pm\)0.51 \\
\bottomrule
\end{tabular}
}

\vspace{3pt}
\end{table*}

\begin{table*}[t]
\centering
\footnotesize
\caption{\textbf{Performance across temporal horizons on \textsc{VSI-Super-Wild}.} The left panel reports \textbf{overall} scores grouped by video duration (minutes). The right panel visualizes per-task scores of four representative models using line charts. (\textbf{Bold}: best, \underline{underline}: second best)}
\label{tab:temporal-horizon}
\vspace{-10pt}
\begin{minipage}[c]{0.42\textwidth}
\setlength{\tabcolsep}{2pt}
\resizebox{0.99\textwidth}{!}{%
\begin{tabular}{lccccc}
\toprule
\textbf{Model/Duration (min)} & \textbf{0–10} & \textbf{10–30} & \textbf{30–60} & \textbf{60–120} & \textbf{120+} \\
\midrule
\rowcolor{proprietaryheader}\multicolumn{6}{l}{\textit{Proprietary Models}} \\
GPT-5.4 & 32.8 & \underline{39.7} & 30.8 & \underline{35.2} & \underline{35.1} \\
Gemini-3.1-Pro & \textbf{51.9} & \textbf{42.6} & \textbf{40.1} & \textbf{46.9} & \textbf{41.2} \\
Gemini-3.1-FlashLite & 26.9 & 24.2 & 21.8 & 23.3 & 20.2 \\
\midrule
\rowcolor{opensourceheader}\multicolumn{6}{l}{\textit{Open-Source Models}} \\
Cambrian-S-0.5B & 23.2 & 21.4 & 21.5 & 20.5 & 14.3 \\
Cambrian-S-3B & 37.3 & 34.4 & 30.1 & 29.7 & 28.7 \\
Cambrian-S-7B & 38.5 & 34.9 & 31.5 & 31.3 & 28.9 \\
Cambrian-S-7B-LFP$^{*}$ & 36.8 & 33.3 & 30.4 & 30.7 & 27.9 \\
Cambrian-S-7B-LFP & 39.4 & 32.7 & 30.0 & 32.2 & 29.7 \\
InternVL3.5-8B & 37.5 & 32.8 & 28.2 & 30.5 & 28.2 \\
Qwen2-VL-7B & 29.5 & 26.0 & 26.0 & 21.3 & 25.7 \\
Qwen2.5-VL-7B & 35.9 & 31.1 & 27.3 & 31.5 & 28.3 \\
Qwen3-VL-8B & 36.6 & 34.7 & \underline{31.7} & 32.6 & 25.9 \\
Qwen3.5-9B & \underline{40.1} & 35.5 & 29.4 & 29.3 & 26.0 \\
Spatial-TTT-nano$^{*}$ & 30.5 & 27.6 & 26.6 & 26.4 & 25.3 \\
\midrule
\textbf{Average} & 35.0 & 31.3 & 28.4 & 28.7 & 26.3 \\
\bottomrule
\end{tabular}
}
\end{minipage}
\begin{minipage}[c]{0.57\textwidth}
    \centering
    \includegraphics[width=\linewidth]{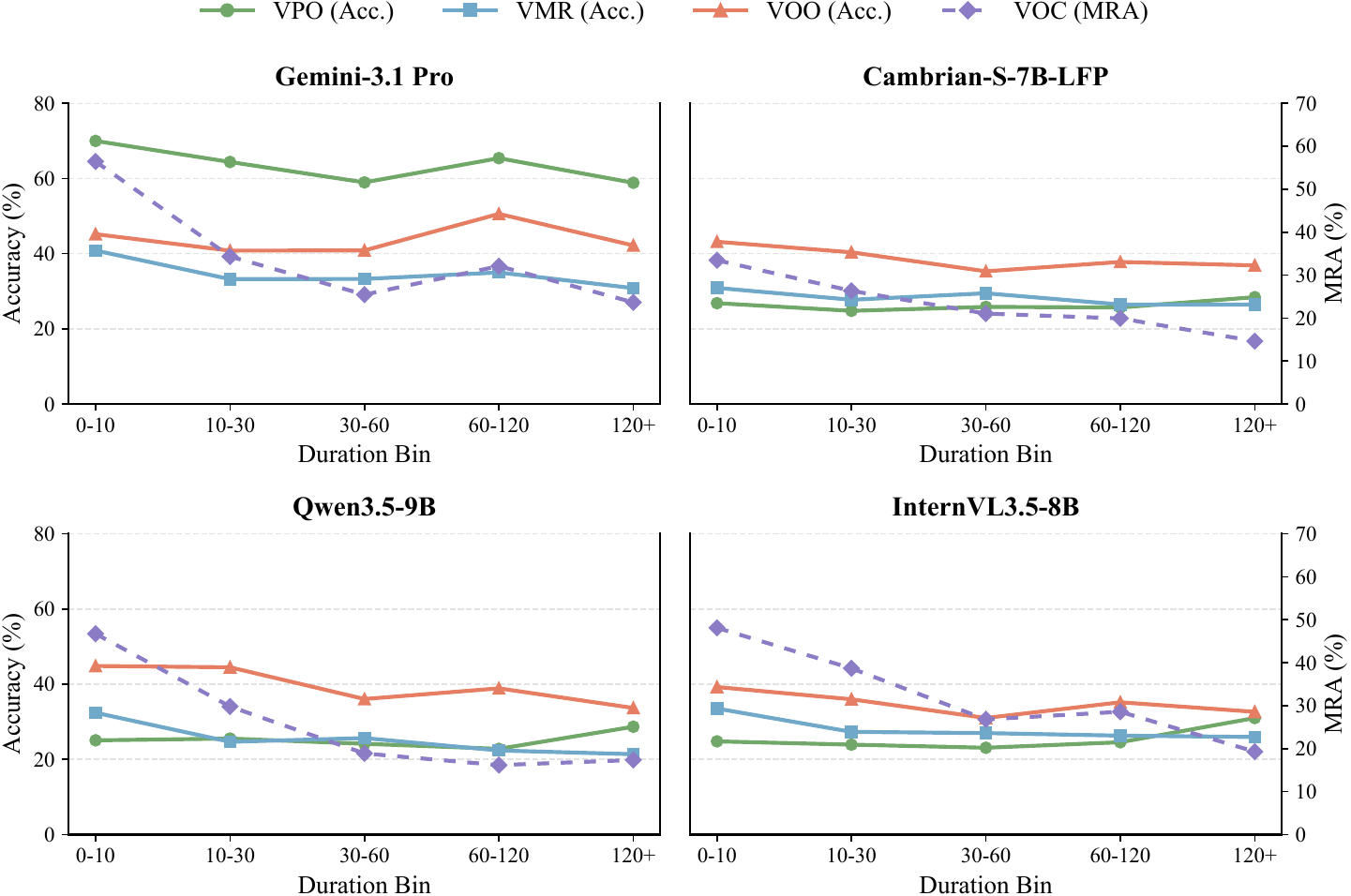}
\end{minipage}
\vspace{-5pt}
\end{table*}

\subsection{Experimental Setup}
\paragraph{Models.} We benchmark 13 representative MLLMs on \textsc{VSI-Super-Wild} to assess their spatial supersensing capabilities in diverse real-world settings. The evaluated models include 3 proprietary MLLMs, namely GPT-5.4~\cite{gpt5}, Gemini-3.1-Pro~\cite{gemini3}, and Gemini-3.1-FlashLite~\cite{gemini3}. We also evaluate 11 open-source MLLMs: Cambrian-S-0.5B/3B/7B(-LFP)~\cite{yang2025cambrians}, Qwen2-VL-7B, Qwen2.5-VL-7B~\cite{bai2025qwen25vltechnicalreport}, Qwen3-VL-8B~\cite{bai2025qwen3vltechnicalreport}, Qwen3.5-9B, InternVL3.5-8B~\cite{wang2025internvl35advancingopensourcemultimodal}, and Spatial-TTT-nano~\cite{spatialttt2026}.

\paragraph{Evaluation Protocol.}
Following the evaluation protocol in \textsc{VSI-Super}~\cite{yang2025cambrians}, we uniformly sample 128 frames at 1080P from each video for all models by default. Models marked with $^{*}$ in Table~\ref{tab:main-exp} are evaluated with streaming inputs instead of the 128-frame sampled input. Query prompts for VMR, VPO, VOO, and VOC are included in the task examples shown in the supplementary material.

\paragraph{Metrics. }
We evaluate all models on four tasks in \textsc{VSI-Super-Wild}: \textbf{VMR}, \textbf{VPO}, \textbf{VOO}, and \textbf{VOC}.
For \textbf{VMR/VPO/VOO}, each query is formulated as a four-way multiple-choice (MC) question, where models select exactly one answer from four candidate options.
We report \emph{Accuracy} (\%), defined as the fraction of queries whose selected option matches the ground-truth answer.
For \textbf{VOC}, each query is formulated as a numeric prediction problem, where models directly output an integer count.
We report \emph{Mean Relative Accuracy (MRA)}, defined as
$\mathrm{MRA}=\frac{1}{N}\sum_{i=1}^{N}\left(1-\frac{| \hat{y}_{i}-y_{i} |}{\max(y_{i},1)}\right)$,
where $y_{i}$ and $\hat{y}_{i}$ denote the ground-truth and predicted counts, respectively.
This metric equals 1.0 for exact matches and decreases smoothly with proportional counting errors.

\subsection{Overall Performance and Key Challenges}
\noindent \paragraph{Overall Performance.}
As shown in Tab.~\ref{tab:main-exp}, current MLLMs still struggle substantially with spatial supersensing in the wild. The strongest model, Gemini-3.1-Pro, reaches an overall score of 44.36, while GPT-5.4 and the strongest open-source model, Cambrian-S-7B, reach 34.52 and 34.22, respectively.
In particular, the numeric task VOC remains challenging even for strong models, with GPT-5.4 reaching 32.94 and the strongest open-source model, InternVL3.5-8B, reaching 36.74. Overall, these results indicate that current MLLMs still struggle to robustly construct, maintain, and query broader spatiotemporal world states under \emph{improved real-world diversity}.

\begin{figure*}[t]
    \centering
    \includegraphics[width=\linewidth]{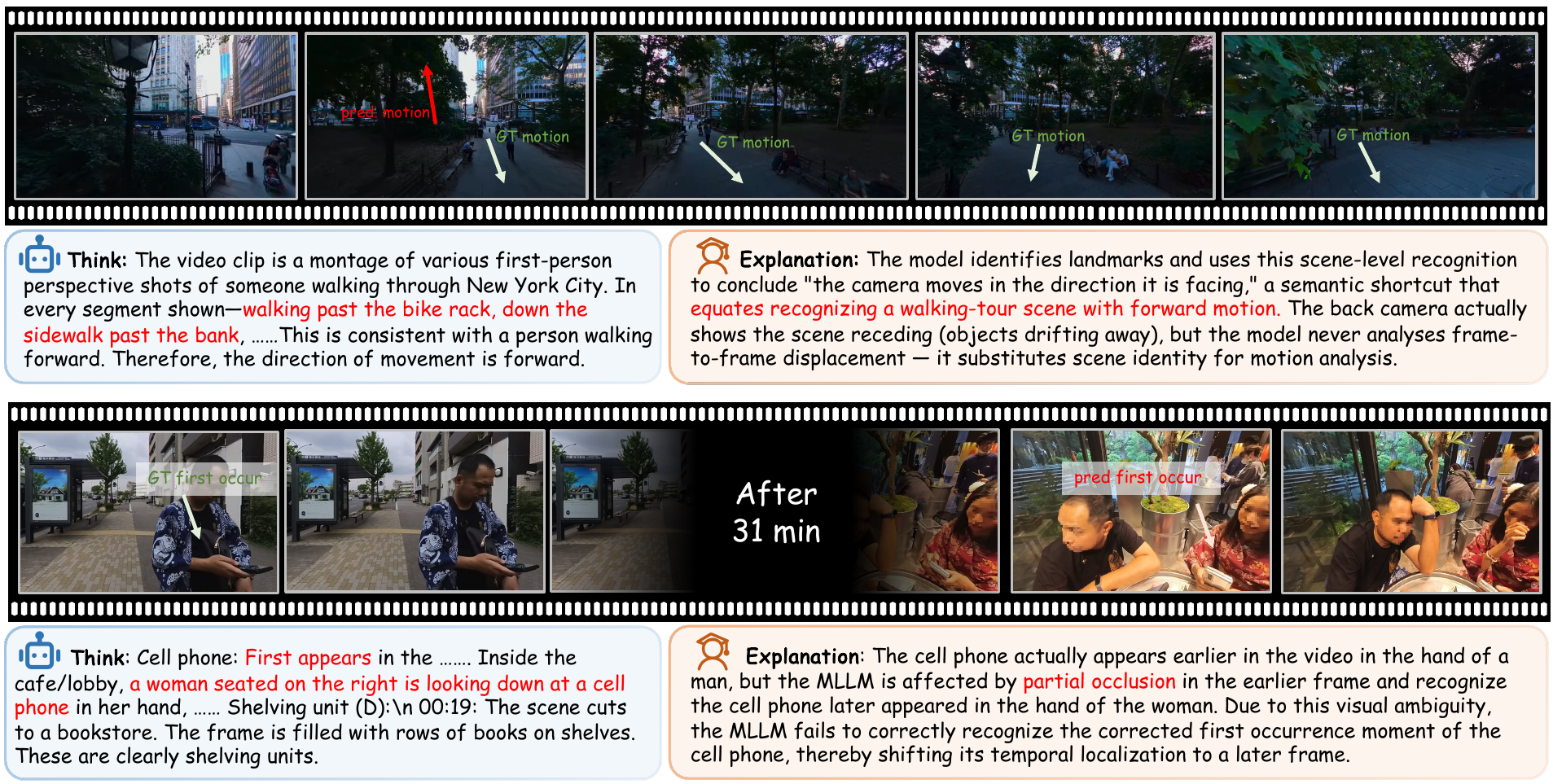}
    \caption{\textbf{Top} (\emph{semantic shortcut}): The model predicts that the observer is moving forward based on frame-level semantic cues (e.g., a typical walking-tour street scene), rather than inferring motion from spatiotemporal evidence across frames. \textbf{Bottom} (\emph{instance confusion}): Under visual challenges such as motion blur and partial occlusion in the wild, the model misses an earlier appearance of the queried object, leading to an incorrect first-occurrence prediction and unstable instance-level tracking over time.}
    \label{fig:error_analysis_si}
\end{figure*}

\begin{figure*}[t]
\centering
\begin{subfigure}{0.47\linewidth}
    \centering
    \includegraphics[width=\linewidth]{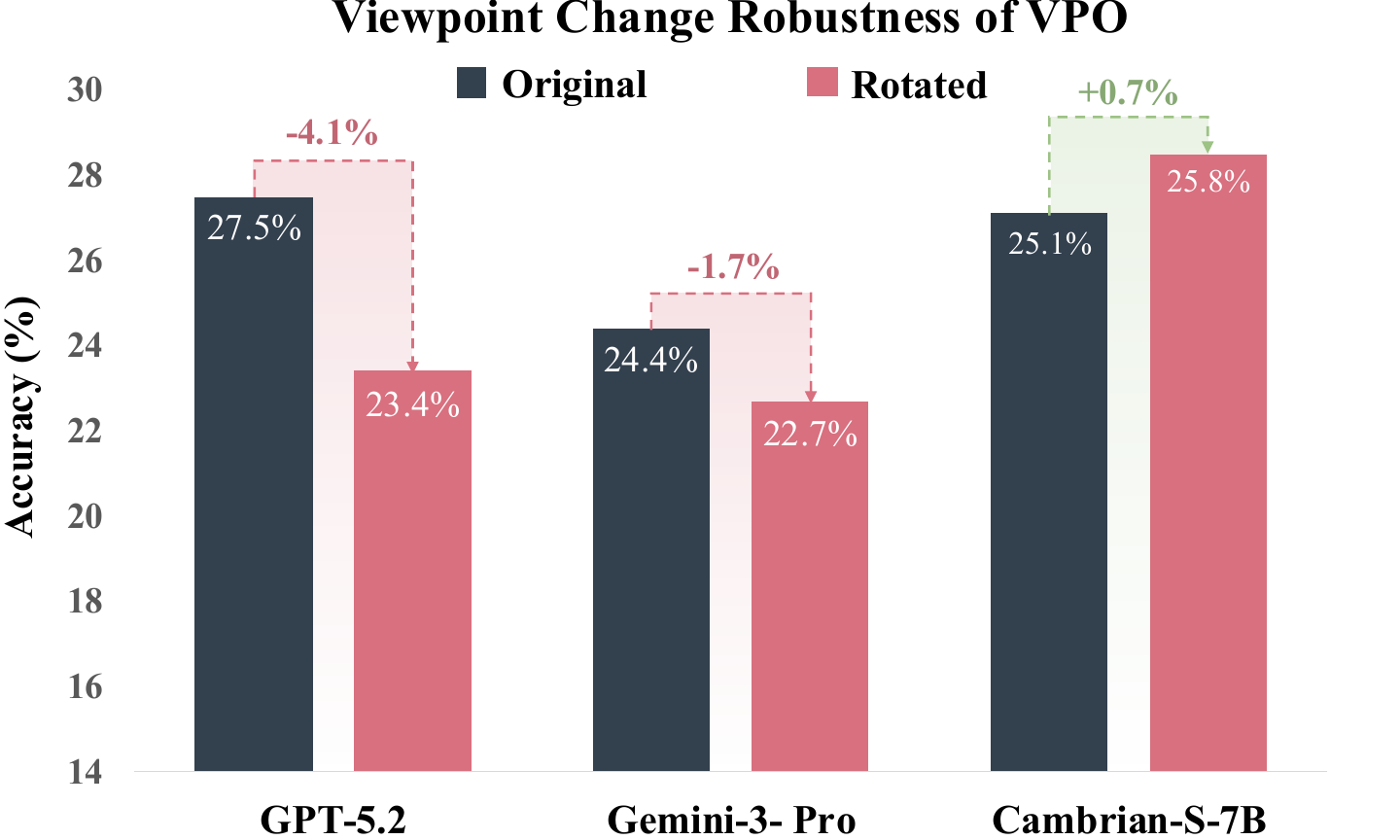}
\end{subfigure}
\hfill
\begin{subfigure}{0.47\linewidth}
    \centering
    \includegraphics[width=\linewidth]{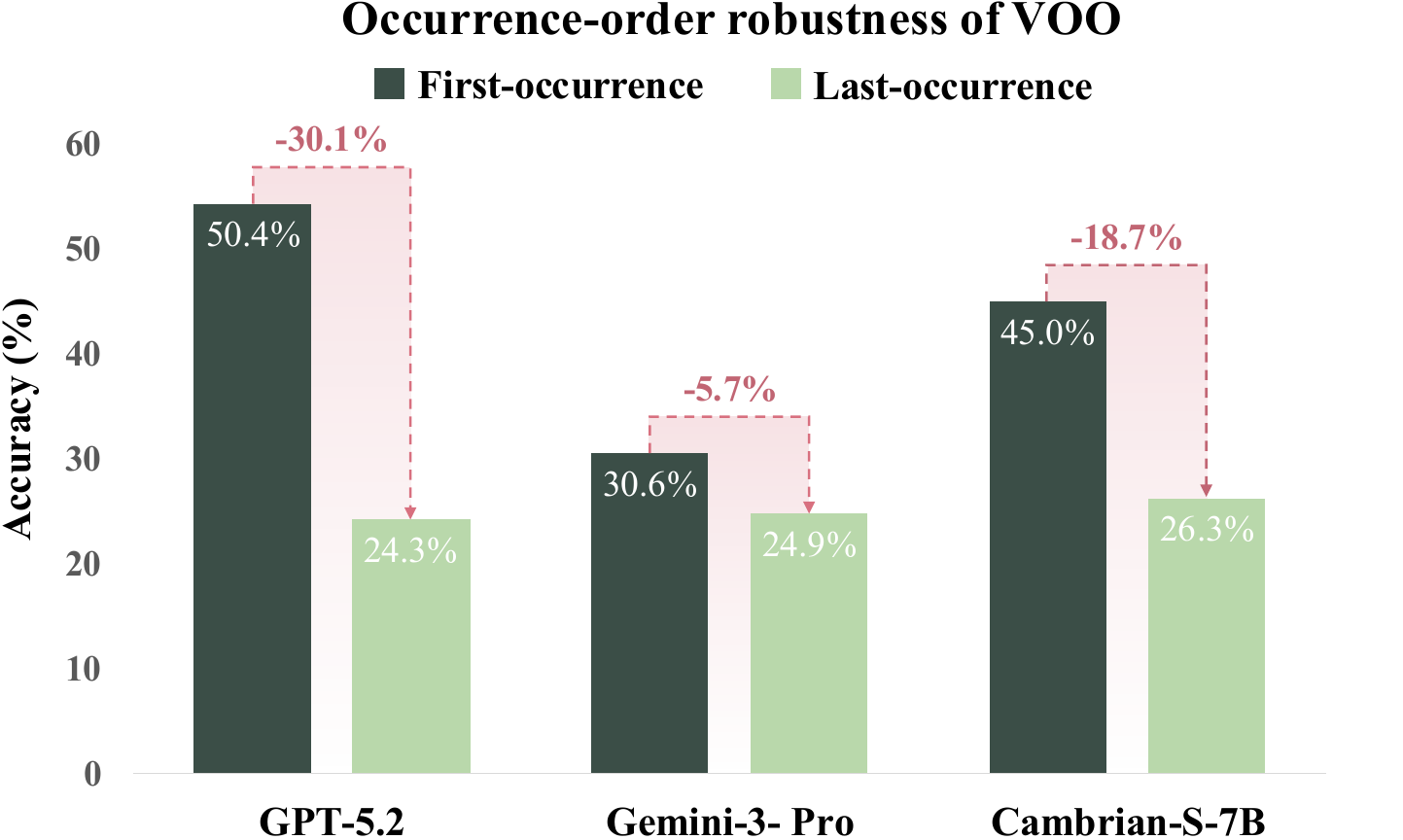}
\end{subfigure}

\caption{\textbf{Left} (\emph{spatial collapse}): We compare VPO accuracy between original and rotated views to test whether models can recognize the same place under viewpoint changes. The drop after rotation suggests that models fail to model a consistent world state after changing views despite the spatially unchanged location. \textbf{Right} (\emph{insufficient update}): We compare VOO accuracy between first- and last-occurrence ordering. The lower performance on last-occurrence queries indicates that models struggle to update world states as later evidence arrives.}
\label{fig:robustness}
\end{figure*}

\noindent \paragraph{Challenges from World State Complexity.}
Beyond overall performance, the results vary systematically across tasks with different world-state anchors.
For a controlled comparison, we focus on VOO, VPO, and VMR, which are all multiple-choice tasks evaluated with the same metric and respectively probe object-, environment-, and agent-centric world states.
As shown in Tab.~\ref{tab:main-exp}, object-state probing is often more tractable for open-source models: for example, Cambrian-S-7B achieves 40.36 on VOO, compared with 26.91 on VMR and 25.58 on VPO, while Qwen3.5-9B reaches 41.25 on VOO but only 25.68 and 25.19 on VMR and VPO.
These results suggest that \emph{agent- and environment-state modeling are systematically harder than object-state modeling} for current MLLMs.
Moreover, prior spatial supersensing benchmarks, which focus on object-centric probing, may underexplore this systematic challenge from world-state complexity.

\noindent \paragraph{Challenges over Longer Temporal Horizons.}
We further analyze how performance varies with temporal horizon. As shown in Tab.~\ref{tab:temporal-horizon}, longer videos are generally more challenging for spatial supersensing: the open-source average score drops from 35.0 in the 0--10 min bin to 26.3 in the 120+ min bin, with the highest average performance occurring in the shortest videos and the lowest in the longest ones. This trend suggests that maintaining coherent world states becomes \emph{increasingly difficult as temporal horizon increases}.
The degradation is not perfectly monotonic for every model, but the overall trend shows that longer video streams lower performance and expose weaknesses in different types of world-state modeling. In particular, the drop across temporal bins suggests that maintaining and updating object-, agent-, and environment-related states becomes harder as the observation horizon expands.

\begin{figure*}[t]
  \centering
  \includegraphics[width=\linewidth]{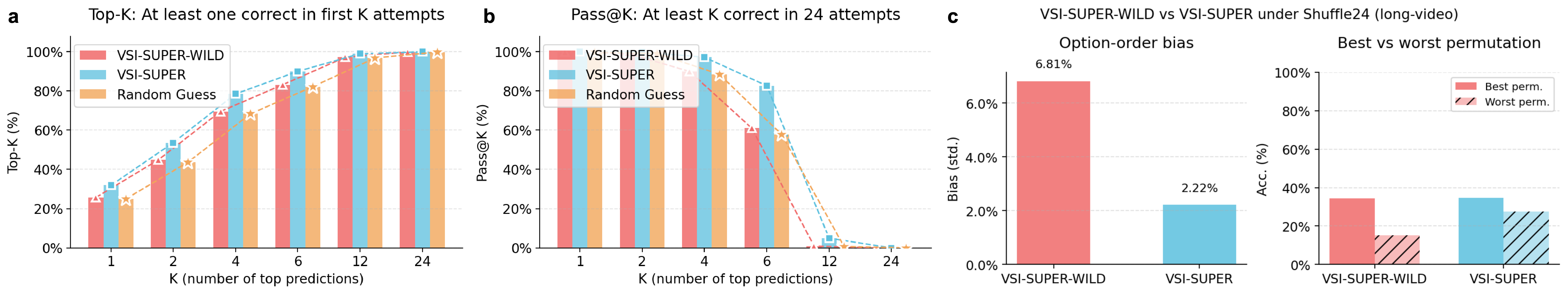}
  \caption{\textbf{VSI-SUPER-WILD vs.\ VSI-SUPER under Shuffle24 (long-video).}
  (\textbf{a}) Top-K: fraction of questions with at least one correct in the first $K$ attempts.
  (\textbf{b}) Pass@K: fraction with at least $K$ correct in 24 attempts.
  (\textbf{c}) Option-order bias: standard deviation of accuracy across 24 option permutations, and best vs.\ worst permutation accuracy.
  VSI-SUPER-WILD shows lower Top-K and Pass@K, higher bias, and a larger best--worst gap than VSI-SUPER, confirming it is more challenging and that Shuffle24 evaluation is essential.}
  \label{fig:shuffle}
\end{figure*}

\subsection{Failure Modes of Spatial Supersensing}

Our results show that current MLLMs perform poorly across \textsc{VSI-Super-Wild}, suggesting that \emph{spatial supersensing in the wild} remains challenging under diverse scenes and longer temporal horizons.
To better understand limitations and provide a roadmap for future research, we conduct both qualitative and quantitative analysis and summarize \emph{four recurring failure modes} that hinder coherent representations of the agent, objects, and environment.

\noindent \paragraph{Spatial Collapse.}
Spatial supersensing requires models to construct a stable spatial world state in the \emph{3D} space, rather than relying on frame-level \emph{2D} semantics.
We examine this through the viewpoint robustness of VPO, which tests whether models can recognize the same place under different viewpoints. Specifically, we compare accuracy between the original and rotated views.
Fig.~\ref{fig:robustness} shows that SOTA models exhibit a clear performance drop after viewpoint change (e.g., GPT-5.4: 26.0\%$\rightarrow$24.9\%; Qwen3-VL-8B: 25.7\%$\rightarrow$24.4\%; Cambrian-S-7B: 26.2\%$\rightarrow$25.6\%), even though the underlying physical location remains unchanged.
Such degradation reflects \emph{spatial collapse}: models fail to maintain a coherent spatial world state across views, and instead fall back to view-specific \emph{2D} frame matching rather than \emph{implicit 3D} world modeling.

\noindent \paragraph{Semantic Shortcut.}
Existing models often rely on frame-level visual semantics instead of continuous spatiotemporal reasoning, using semantic cues from individual frames as shortcuts for world-state inference.
We illustrate this failure mode with VMR, where motion orientation requires constructing an agent-centric state from the video stream, yet models can still be misled by local semantic cues and produce plausible-but-incorrect answers.
As shown in the top of Fig.~\ref{fig:error_analysis_si}, the agent is actually moving backward relative to its viewing direction, but the model predicts ``forward'' because the straight road ahead provides a strong semantic cue.
This failure suggests that models \emph{exploit semantic shortcuts} instead of inferring motion from spatiotemporal evidence and updating an agent-centric world state.

\noindent \paragraph{Insufficient Update.}
Beyond initial recognition, spatial supersensing requires continuously updating world states as video streams arrive.
We probe this using VOO first- vs.\ last-occurrence ordering (Fig.~\ref{fig:robustness}), where first-occurrence mainly tests registering objects at their initial encounter, while last-occurrence requires updating world states to reflect later evidence.
Across models, accuracy is substantially higher for first-occurrence than for last-occurrence ordering (e.g., GPT-5.4: 31.3\% vs.\ 22.1\%; Qwen3-VL-8B: 50.2\% vs.\ 30.5\%; Cambrian-S-7B: 50.7\% vs.\ 30.0\%).
This gap suggests that models can preserve early world states relatively well, but struggle to \emph{sufficiently update world states} as new evidence arrives.

\noindent \paragraph{Instance Confusion.}
For spatial supersensing in the wild, models need to robustly recognize objects under challenging real-world conditions, such as motion blur, partial occlusion, and viewpoint change.
However, as shown in the bottom of Fig.~\ref{fig:error_analysis_si}, the model misses an earlier appearance of the queried object under blur and partial occlusion, leading to incorrect ordering.
This indicates weak object-level identity tracking and unstable object-centric state modeling under in-the-wild visual noise and dynamics.
Overall, these failures underscore that \emph{in-the-wild} visual complexity remains highly challenging for spatial supersensing.

\subsection{Long-Video Benchmark: VSI-SUPER-WILD vs.\ VSI-SUPER under Shuffle24}
To examine whether VSI-SUPER-WILD introduces additional difficulty beyond existing long-video benchmarks, we compare it with VSI-SUPER under \textbf{Shuffle24}. For each multiple-choice question, we evaluate all 24 permutations of the four answer options and report metrics that are invariant to option order. This protocol reduces dependence on a specific option arrangement and reveals whether model predictions are stable under semantically equivalent choices.

Fig.~\ref{fig:shuffle} reports four complementary views. \textbf{Top-K} measures the fraction of questions with at least one correct prediction in the first $K$ attempts, while \textbf{Pass@K} measures the fraction with at least $K$ correct predictions among all 24 permutations. For the same $K$, \textsc{VSI-Super-Wild} is consistently lower than \textsc{VSI-Super} on both metrics and remains closer to the random-guess baseline, indicating greater difficulty in recovering the correct answer even across shuffled trials. We analyze option-order sensitivity: Fig.~\ref{fig:shuffle}(c) shows higher \textbf{option-order bias} on \textsc{VSI-Super-Wild}, measured by the standard deviation of accuracy across permutations, and Fig.~\ref{fig:shuffle}(d) shows a larger gap between the best and worst permutations. Overall, VSI-SUPER-WILD is more challenging and less stable under option shuffling, making Shuffle24 necessary to avoid overestimated performance.

\section{Conclusion}
To assess whether MLLMs can conduct human-like implicit world modeling under real-world in-the-wild streams, we introduce \textbf{\textsc{VSI-Super-Wild}}, a large-scale benchmark for evaluating \emph{spatial supersensing in the wild}. Compared with existing supersensing benchmarks, \textsc{VSI-Super-Wild} advances evaluation along two axes: \emph{real-world diversity} and \emph{broader world-state probing}. For real-world diversity, we curate 442 unconstrained online videos, spanning 8 scene categories and lasting over 4 hours at maximum, without generative editing or clip concatenation. For broader world-state probing, we construct 6,980 human-verified Q\&As across four task types, forming a cognitively grounded task suite that probes object-, agent-, and environment-centric components of world states.

We benchmark 13 mainstream MLLMs on \textsc{VSI-Super-Wild} and find that current models still fall substantially short of robust spatial supersensing in the wild. A closer look reveals clear \emph{performance disparities} across supersensing conditions: models are relatively more reliable on object-centric state queries, yet struggle on agent- and environment-centric components that require stronger 3D spatial cognition; tasks that demand precise numeric answers under continual state updates, such as continuous counting, are especially challenging. Moreover, performance degrades markedly as the temporal horizon increases, suggesting difficulty in sustaining consistent world-state updates over extended streams. Qualitative error analyses further point to recurring failure patterns—spatial collapse, semantic shortcuts, insufficient update, and instance confusion—which help explain these disparities and indicate that current MLLMs still over-rely on frame-level semantics rather than maintaining stable, queryable spatiotemporal world representations. Together, these findings highlight the key bottlenecks that future models must address to achieve reliable spatial supersensing in the wild.

{
    \small
    \bibliographystyle{ieeenat_fullname}
    \bibliography{main}
}

\clearpage

\appendix

\title{Supplementary Material of \textbf{\textsc{VSI-Super-Wild}}}

\maketitle

\begin{abstract}
This appendix provides additional implementation details, experimental results, and supplementary analyses that further support the methodology, evaluation, and findings presented in the main paper.
\begin{itemize}
    \item §~\ref{sec:data_construction} details the semi-automatic data construction pipeline including video collection, metadata generation, question-answer generation with human-in-the-loop verification.
    \item §~\ref{sec:annotation} describes the manual metadata verification interface and workflow  used to validate and complete the dataset.
    \item §~\ref{sec:task_examples} presents detailed examples and visualizations of all proposed tasks (VMR, VPO, VOO, and VOC) in \textsc{VSI-Super-Wild}.
\end{itemize}

\end{abstract}

\section{Data Construction}
\label{sec:data_construction}

This section provides additional implementation details for the data construction process described in the main paper. We further elaborate on the semi-automatic data construction pipeline with human-in-the-loop verification used to build our benchmark dataset.
The pipeline consists of three stages: (1) \textbf{Video Collection}, (2) \textbf{Metadata Generation}, and (3) \textbf{Question-Answer Generation}. The detailed procedures are described below.

\subsection{Video Collection}
\subsubsection{Online video crawling.}
To construct a large and diverse collection of panoramic videos, we crawl videos from YouTube using a keyword-based retrieval strategy. 
Each query is composed of two parts: (i) a \emph{panorama cue} (e.g., $360^{\circ}$, panorama) and (ii) a \emph{category cue} indicating one of eight video categories (culture \& entertainment, industry, medical, office \& education, residential space, retail space, street view, and transportation hubs).
To improve coverage, we further leverage a multimodal large language model (MLLM) to expand these seed keywords into additional variants and multilingual forms, which are then used for large-scale crawling on YouTube.

\subsubsection{Manual video filtering.}
To ensure data quality, we manually remove low-quality videos and videos captured with static cameras or mounted platforms (e.g., in-car or on-ship recordings). This filtering step keeps predominantly egocentric, dynamic, and handheld videos that better match our target setting.

\subsubsection{Panorama-to-perspective projection.}
Panoramic videos are typically stored in an equirectangular format. We therefore project each panoramic frame into four orthogonal perspective views (\emph{Front, Back, Left, Right}) with a field of view of $90^{\circ}$ and an output resolution of $1920 \times 1080$. For privacy protection, we further apply SCRFD-10G to each projected video to detect human faces and blur all detected face regions \cite{scrfd}. We treat the resulting four perspective streams as independent videos and add them to the data pool for subsequent processing. Unless otherwise specified, the term ``video'' in the remainder of this paper refers to these projected perspective videos.

\subsection{Metadata Generation}
\subsubsection{Temporal Metadata Generation}
We generate object-centric temporal metadata that records fine-grained objects appearance in each videos. This metadata supports \textbf{object order} and \textbf{object counting} tasks, and is represented as instance masks associated with timestamps for each selected object.

\emph{Candidate object proposal.}
We start by asking human experts to curate a seed list of task-friendly objects that are semantically unambiguous, countable, non-abstract, and easy to recognize. Following the same criteria, we additionally employ an MLLM (Gemini-3-Pro) to propose video-specific candidates based on 16 keyframes per video, improving diversity and adapting the vocabulary to the scene content. To reduce redundancy between the human and MLLM proposals, we normalize candidates using CLIP-text embedding similarity with a threshold of $0.92$; when a human-proposed candidate is deemed synonymous with an MLLM-proposed one, we keep the MLLM term as the canonical label.

\emph{Open-vocabulary detection and frequency-based filtering.}
To ensure that candidates are detectable and suitable for object-centric tasks, we run open-vocabulary detection using {YOLO-World-v2} on videos that are uniformly downsampled to 2 FPS. We use candidate concepts as textual prompts and compute temporal occurrence statistics for each concept. We then select \emph{task-friendly objects} by removing concepts with extremely low frequency (insufficient evidence) or extremely high frequency (overly redundant), which helps control task difficulty.

\emph{Instance segmentation and metadata construction.}
For the retained objects, we apply Segment Anything Model 3 ({SAM3}) to obtain instance-level masks. Each annotation is indexed by a concept label and an instance identifier, and is associated with a binary mask for the corresponding object instance. Due to GPU memory constraints, we downsample videos to 2 FPS and split them into non-overlapping 5-minute clips, which are processed independently by SAM3. We then perform manual de-duplication to reconcile repeated instances across clips. The resulting annotations form fine-grained temporal metadata, consisting of instance masks paired with their corresponding timestamps.

\subsubsection{Spatial Metadata Generation}
We generate spatial metadata by estimating the camera pose at each timestamp throughout the video. In detail, we apply Depth-Anything-3 ({DA3}) to videos downsampled to 2 FPS to obtain time-varying camera poses. This signal supports the motion orientation recall task, and the results are subsequently audited through manual review.

\subsection{Question-Answer Generation}
\subsubsection{From metadata to QA pairs.}
Given the temporal and spatial metadata together with the associated videos, we synthesize final question-answer (QA) pairs for four tasks: \textbf{VMR}, \textbf{VPO}, \textbf{VOO}, \textbf{VOC}. These tasks adopt different answer formats, including multiple-choice (MC) and numeric responses. The rule-based QA generation procedure for each task are as follows.

\begin{itemize}
    \item \textbf{VMR (MC).} For each video, we sample a set of timestamps at which the camera exhibits notable motion (one every 5 minutes) based on the spatial metadata. We then construct multiple-choice questions whose options are directional descriptions derived from the estimated camera motion and orientation.
    
    \item \textbf{VPO (MC).} For each video, we sample four anchor frames and, at each anchor timestamp, generate perturbed views by rotating the corresponding panorama to obtain nearby perspective observations while ensuring sufficient overlap. We assign indices to these views and construct multiple-choice questions that ask for the correct order among them.
    
    \item \textbf{VOO (MC).} For each video, we compute the first-appearance and last-appearance orders of candidate objects from the temporal metadata, and generate corresponding multiple-choice QA pairs.
    
    \item \textbf{VOC (Numeric).} For each video, we generate counting questions directly from the temporal metadata under a streaming setting: the same counting query is issued at fixed intervals (every 5 minutes), and the answer is obtained from metadata-derived counts.
\end{itemize}

\subsubsection{Human verification.}
We incorporate human verification during QA generation to ensure the validity of QA pairs (see details in Sec.~\ref{sec:annotation}). Human experts review metadata-QA consistency and correct errors such as duplicated or missed object counts arising from inaccurate temporal metadata, pose drift caused by erroneous spatial metadata, and questions that are ambiguous or not answerable from the video evidence. When necessary, the pipeline rolls back to refine the underlying metadata and re-generate the corresponding QA pairs.

\section{Verification Interface and Workflow}
\label{sec:annotation}

This section provides additional details of the human verification step introduced in Appendix~\ref{sec:data_construction}. Rather than annotating data from scratch, annotators use dedicated verification interfaces to review, validate, and revise automatically generated temporal and spatial metadata produced by the semi-automatic pipeline. We organize the verification process into two parts: verification for temporal metadata and verification for spatial metadata. The former is used for VOO and VOC that depend on object occurrences and temporal accumulation, while the latter is used for VMR that depend on agent motion modeling.

\subsection{Verification for Temporal Metadata}

We use temporal metadata verification to validate automatically generated metadata for VOO and VOC. This metadata includes the first and last appearance order of queried objects, the temporal span of each object instance, and the clip-wise unique count of each object. It is further used to derive question--answer pairs for VOO and VOC tasks. Annotators verify these generated results against the video content and revise them when necessary, rather than annotating samples from scratch.

\begin{figure*}[t]
    \centering
    \includegraphics[width=0.9\linewidth]{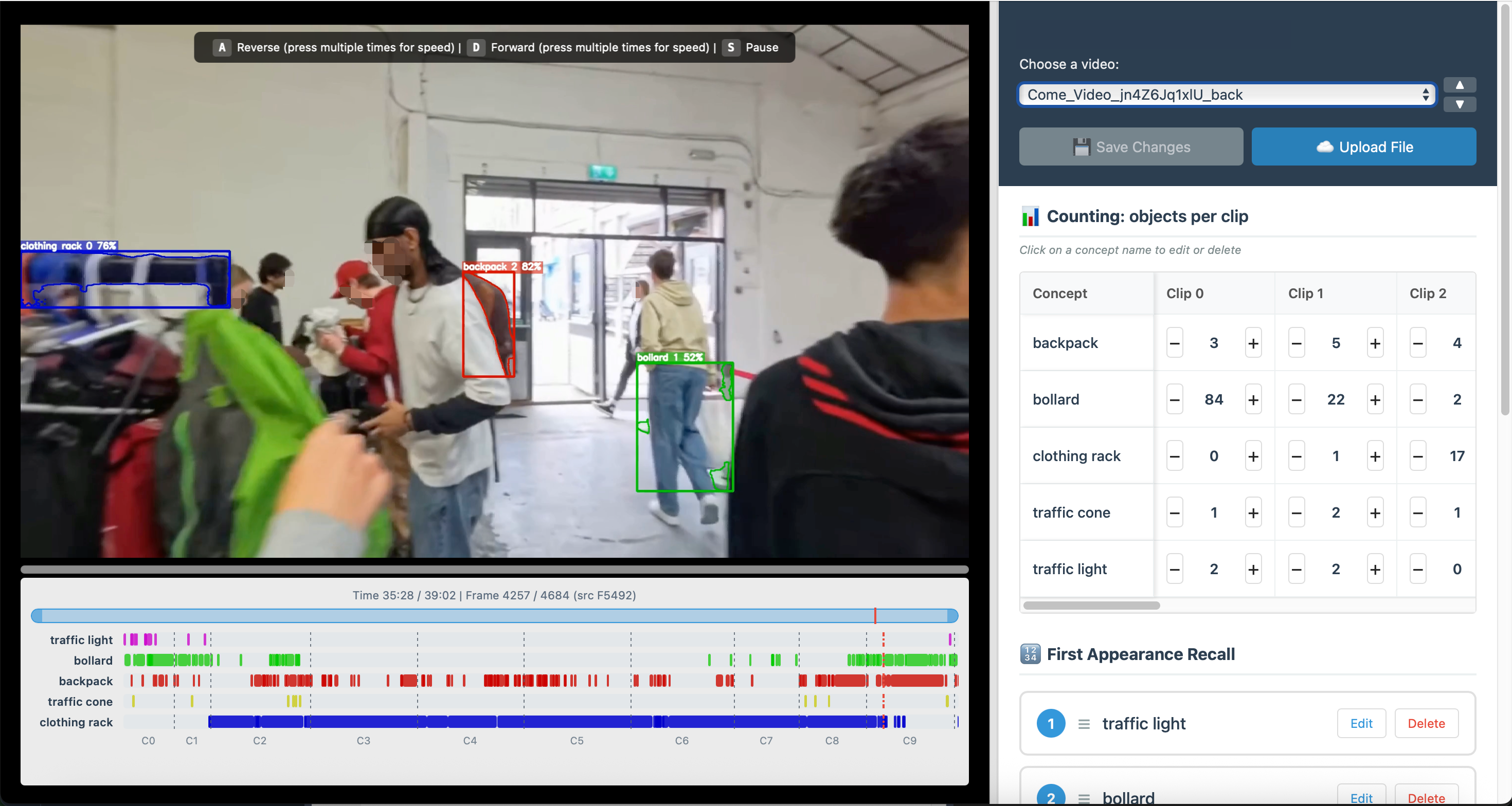}
    \caption{\textbf{Verification interface for temporal metadata} used in VOO and VOC. The interface displays SAM3-generated object masks, timeline visualizations of object instances, and an appearance-order adjustment module, allowing annotators to revise automatically generated temporal metadata rather than annotate samples from scratch.}
    \label{fig:interface-v3}
\end{figure*}

\paragraph{Verification Interface.} As shown in Fig.~\ref{fig:interface-v3}, the temporal verification interface contains a video verification area and a control panel. In the video verification area, the upper panel displays the video with SAM3-generated object masks overlaid on the frames, allowing annotators to directly inspect the quality of the generated object annotations. The lower panel provides a timeline visualization of object instances, showing their temporal spans in the video and the positions of corresponding frame segments. This timeline helps annotators verify whether the generated metadata correctly captures when an object first appears, how long it remains visible, and whether repeated detections correspond to the same instance. The control panel further includes an appearance-order adjustment module, which lists objects according to their first and last appearances and allows annotators to revise the generated ordering when necessary.

\paragraph{Verification Workflow.} During verification, annotators first select a video sample from the control panel and inspect the SAM3-generated masks over the video. They then verify whether the generated temporal metadata correctly captures the relevant object instances, including their temporal spans, first appearances, and clip-wise unique counts. For appearance order, annotators check whether the generated metadata matches the order in which target objects first appear in the video. For clip-wise object counts, they verify whether the metadata correctly records the cumulative number of unique objects observed up to each clip. When inconsistencies are found, annotators revise the generated metadata through the interface rather than re-annotating the entire sample.

\paragraph{Verification Criteria.} To ensure that the verified temporal metadata is reproducible and reliable for downstream evaluation, annotators follow several criteria during verification. First, incorrect object categories produced by SAM3 are corrected by editting the concept name when the generated labels do not match the visual content. Second, duplicated detections of the same object across adjacent or overlapping temporal segments are merged so that each valid instance is counted only once. Third, if a target object appears in the video but no corresponding mask is generated, the missing object is manually supplemented in the counting result. Finally, if the generated appearance order is inconsistent with the actual video content, the ordering is recalibrated through the appearance-order module. Through this process, temporal metadata verification ensures that object-level state information remains consistent with the observed video stream.

\subsection{Verification for Spatial Metadata}
\label{sec:spatial_metadata_verification}

We use spatial metadata verification to validate automatically generated metadata for VMR. This metadata includes the pivot frame, the ego-motion direction of the camera, and the corresponding orientation angle for each clip. It is then used to derive question--answer pairs for motion reasoning. Annotators verify these generated results against the video content and revise them when necessary, rather than performing manual annotation from scratch.

\begin{figure*}[t]
    \centering
    \includegraphics[width=0.9\linewidth]{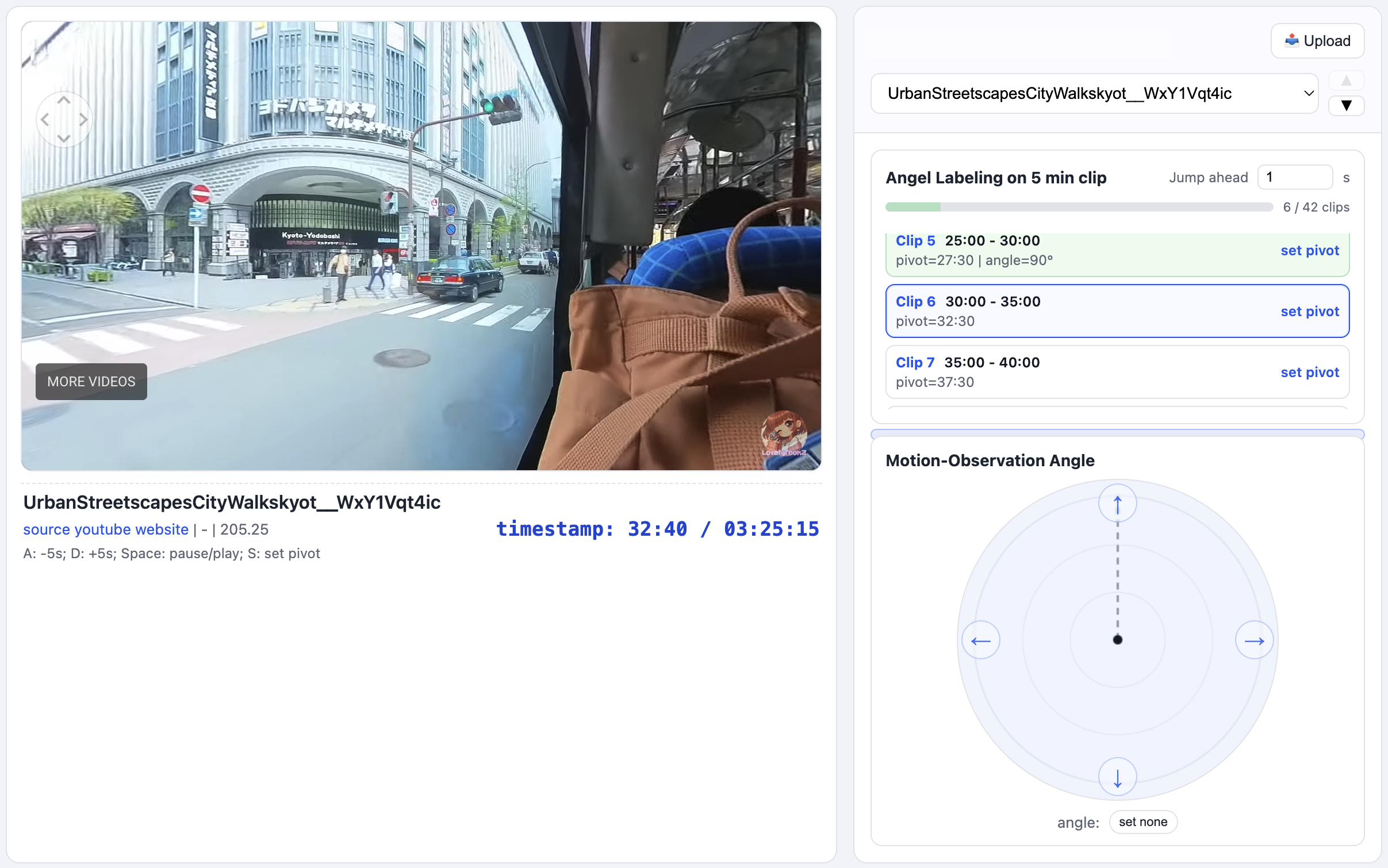}
    \caption{\textbf{Verification interface for spatial metadata} used in VMR. The interface presents the video, clip-level metadata, and a compass-like orientation tool, enabling annotators to verify and revise generated motion directions through a human-in-the-loop quality-control process.}
    \label{fig:VMR-ui}
\end{figure*}

\paragraph{Verification Interface.} As shown in Fig.~\ref{fig:VMR-ui}, the spatial verification interface also consists of a video verification area and a control panel. The video verification area contains a playback window for inspecting the scene and the camera motion, together with auxiliary metadata such as the video source, current timestamp, and total duration. It also provides keyboard shortcuts for efficient verification, including skipping forward or backward, pausing playback, and setting a pivot frame. The control panel contains three modules. The clip-management module allows annotators to select videos and monitor verification progress within each 5-minute segment. The clip-list module displays the start and end time of each clip together with its pivot timestamp and motion angle. The orientation module provides a compass-like interface in which annotators can adjust a directional vector to verify or revise the generated motion direction, or mark the clip as having no valid motion when appropriate.

\paragraph{Verification Workflow.} During verification, annotators first select a video and inspect each clip in the corresponding 5-minute segment. For each clip, they identify a pivot frame at which the motion direction is visually stable and easy to determine. They then check whether the generated spatial metadata matches the observed camera motion and revise the direction through the orientation tool when necessary. If no suitable pivot frame can be identified within a clip, that clip is skipped and no question-answer pair is generated from it.

\paragraph{Verification Criteria.} To ensure consistency and usability of the verified spatial metadata, annotators follow several criteria. First, the pivot frame should correspond to a moment at which the motion direction is stable and representative of the clip. Second, the verified angle should accurately capture the instantaneous heading of the camera relative to the scene. Third, if no meaningful motion is present in the clip, the direction should be marked as none rather than forcing an angle label. Finally, motion directions across adjacent clips should remain temporally coherent and physically plausible. These criteria help ensure that the verified spatial metadata faithfully reflects the motion cues required for VMR.

Overall, the verification interface and workflow provide an efficient human-in-the-loop quality-control mechanism for the semi-automatic pipeline described in Appendix~\ref{sec:data_construction}. By separating temporal metadata verification from spatial metadata verification while maintaining a consistent verification-based framework, we tailor human verification to the distinct requirements of different task families without introducing a separate annotation stage.

\section{Task Examples and Visualizations}
\label{sec:task_examples}

\begin{figure*}
    \centering
    \includegraphics[width=0.95\linewidth]{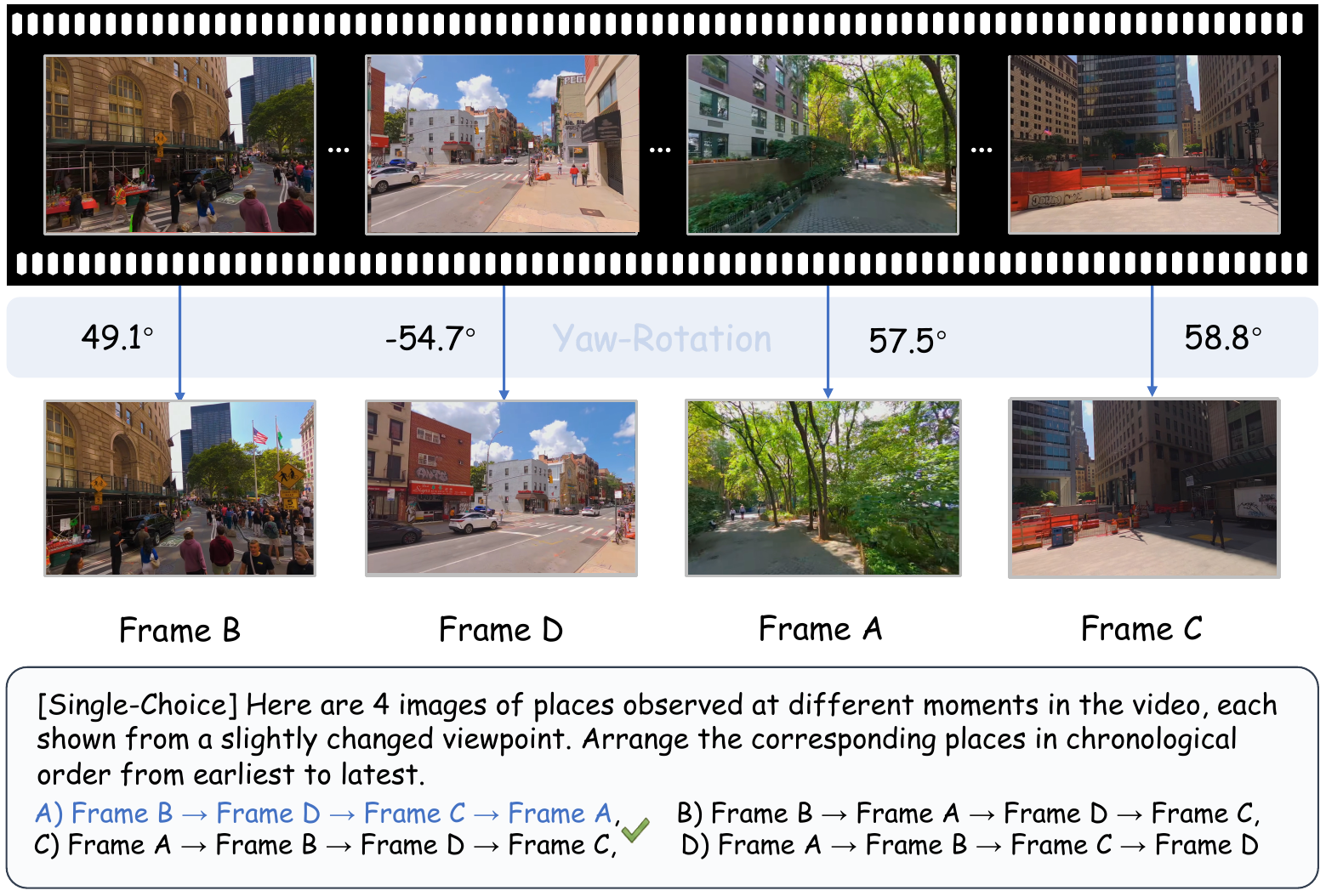}
    \caption{
Example of the \textbf{VPO} task.
Four frames captured at different moments in the video are presented with
yaw-rotated viewpoints (30°–60°). The model must determine the
correct chronological order by reasoning about spatial layout and camera
movement across the video.
}

    \label{fig:eg_VPO}
\end{figure*}

\begin{figure*}
    \centering
    \includegraphics[width=0.95\linewidth]{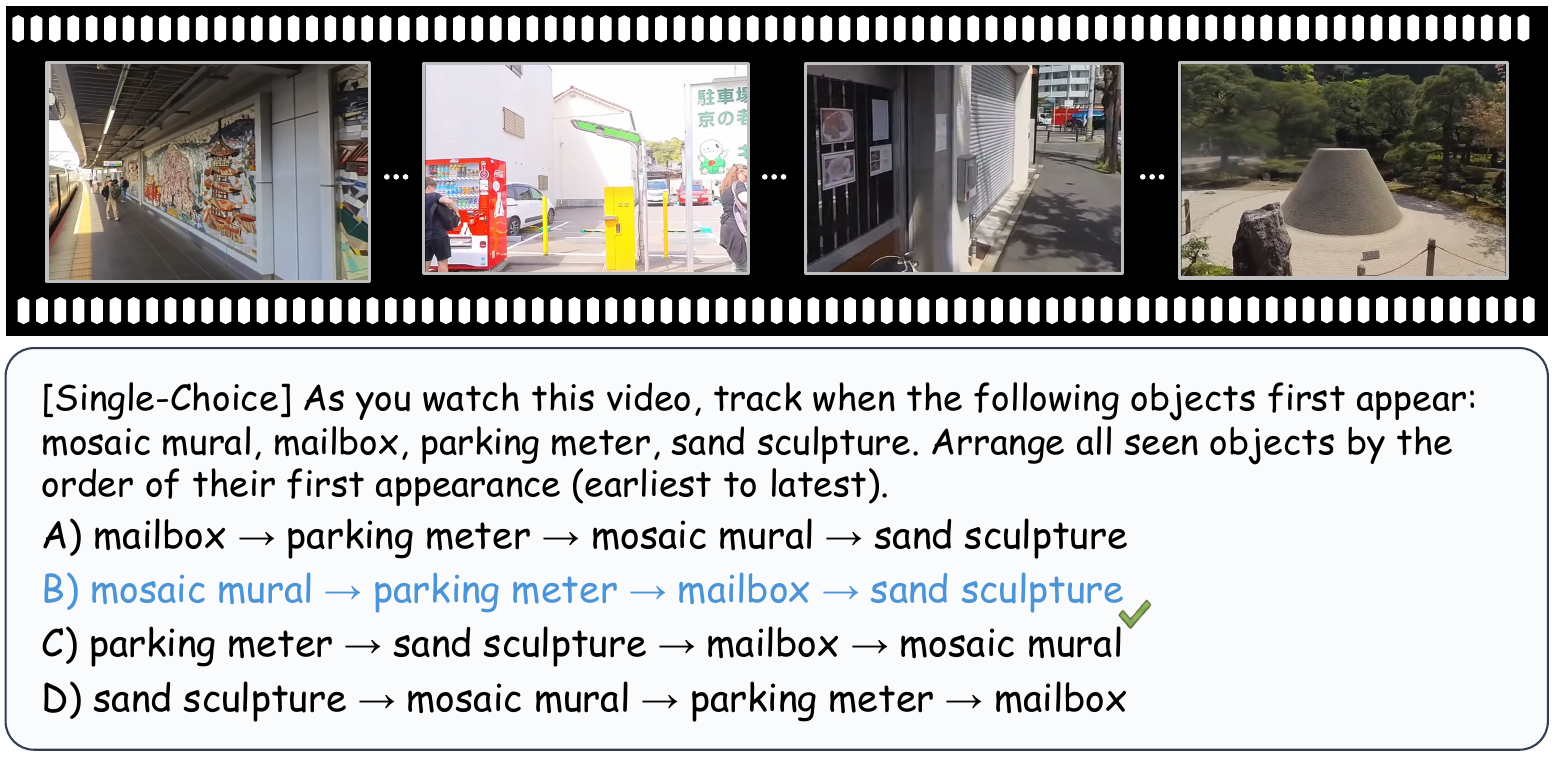}
    \caption{
Example of the \textbf{VOO} task.
The model must track multiple objects throughout the video and determine
the order in which they first appear. This task requires maintaining
long-term memory of object occurrences across the video stream.
}

    \label{fig:eg_VOO}
\end{figure*}

\begin{figure*}
    \centering
    \includegraphics[width=0.95\linewidth]{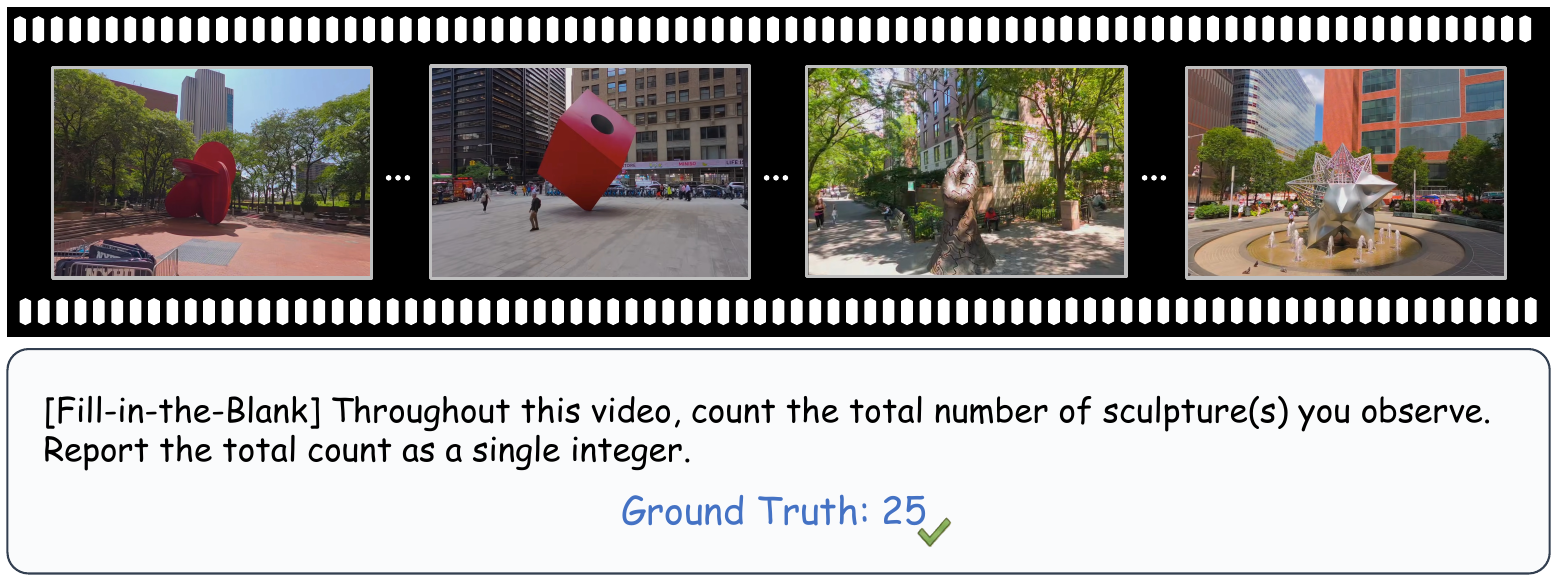}
    \caption{
Example of the \textbf{VOC} task.
The model observes a full video stream and must count the total number of
target objects appearing throughout the video. Accurate counting requires
maintaining temporal consistency and avoiding double-counting across
multiple frames.
}

    \label{fig:eg_VOC}
\end{figure*}

To provide an intuitive understanding of our four tasks centered on the agent, object, and environment, Figures~\ref{fig:eg_VPO}--\ref{fig:eg_VOC} present representative query examples and visual frames for VMR, VPO, VOO, and VOC. These tasks probe complementary aspects of spatial supersensing in real-world video streams. \textbf{VMR} focuses on agent-centric reasoning by requiring the model to infer camera ego-motion from temporal visual evidence. \textbf{VPO} focuses on environment-level reasoning by asking the model to recover the temporal order of queried places despite viewpoint changes. \textbf{VOO} focuses on object-centric temporal reasoning by requiring the model to identify the order in which target objects first appear. \textbf{VOC} further extends object-level reasoning to cumulative counting over time, requiring consistent object-state maintenance across the full stream.

\begin{figure*}
    \centering
    \includegraphics[width=0.95\linewidth]{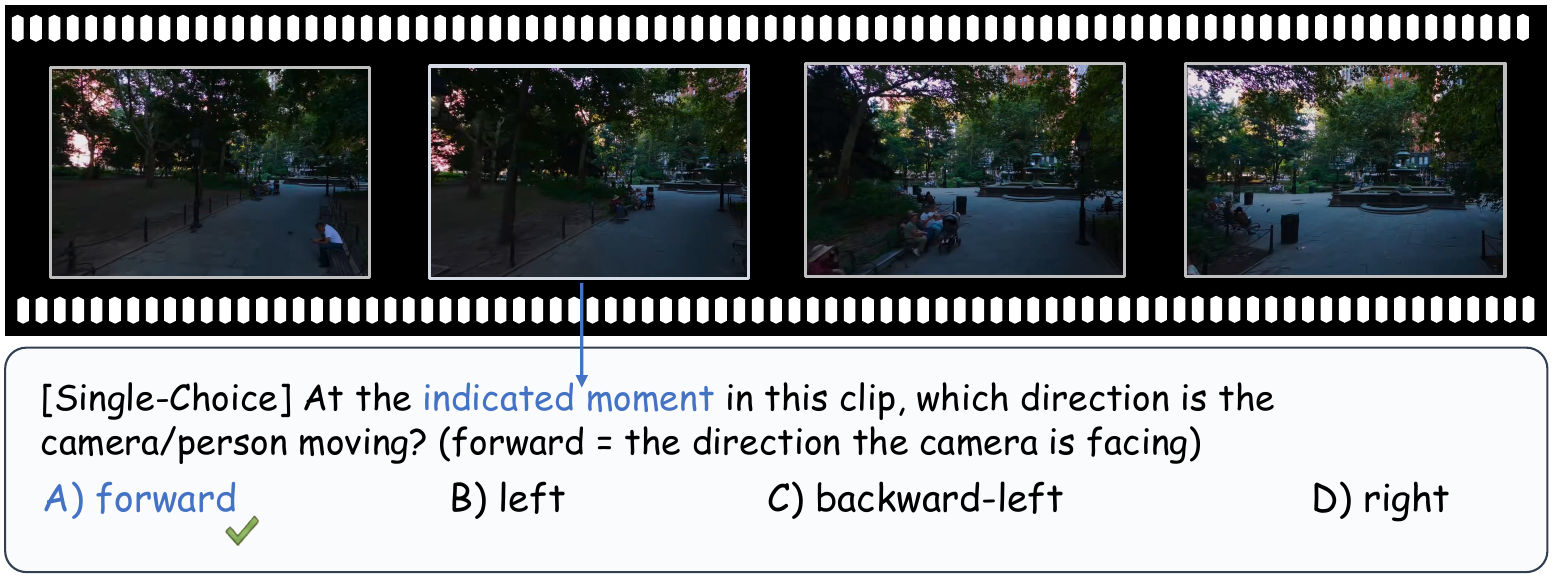}
    \caption{
Example of the \textbf{VMR} task.
Given a frame from a first-person video stream, the model must infer the
ego-motion direction of the camera relative to its viewing orientation.
The correct answer is determined by analyzing perspective changes and
object displacement across the temporal context.
}

    \label{fig:eg_VMR}
\end{figure*}

\end{document}